\documentclass[10pt,twocolumn,letterpaper]{article}

 \usepackage{cvpr}              

\usepackage[dvipsnames]{xcolor}

\definecolor{cvprblue}{rgb}{0.21,0.49,0.74}
\usepackage[pagebackref,breaklinks,colorlinks,citecolor=cvprblue]{hyperref}

\newlength\savewidth\newcommand\shline{\noalign{\global\savewidth\arrayrulewidth
    \global\arrayrulewidth 1pt}\hline\noalign{\global\arrayrulewidth\savewidth}}
    \newcommand{\tablestyle}[2]{\setlength{\tabcolsep}{#1}\renewcommand{\arraystretch}{#2}\centering\footnotesize}
\usepackage{array}
\newcolumntype{Y}[1]{>{\centering\arraybackslash}p{#1}}
\usepackage{times}
\usepackage{epsfig}
\usepackage{graphicx}
\usepackage{amsmath}
\usepackage{wrapfig,lipsum,booktabs}
\usepackage[dvipsnames]{xcolor}
\usepackage{mathtools} 
\usepackage{multirow}
\usepackage{graphicx}  
\usepackage{amssymb}
\usepackage{color}
\usepackage{tabulary}

\usepackage{float}
\usepackage{gensymb}
\newcommand{\cmark}{\ding{51}}
\newcommand{\xmark}{\ding{55}}
\usepackage{colortbl}
\usepackage{comment}
\makeatletter
\def\hlinewd#1{%
\noalign{\ifnum0=`}\fi\hrule \@height #1 \futurelet
\reserved@a\@xhline}

\usepackage{pifont}

\definecolor{recolor}{rgb}{0,1,0}

\definecolor{srcolor}{rgb}{1,0,0}

\definecolor{shcolor}{rgb}{0,0,1}

\def\paperID{2792} 
\def\confName{CVPR}
\def\confYear{2024}
\title{Unifying Correspondence, Pose and NeRF for Generalized Pose-Free Novel View Synthesis}

\author{
Sunghwan Hong\\
Korea University \\
\and
Jaewoo Jung\\
Korea University \\
\and
Heeseong Shin\\
Korea University \\
\and
Jiaolong Yang \\
Microsoft Research Asia\\
\and
Seungryong Kim \\
Korea University \\
\and
Chong Luo \\
Microsoft Research Asia \\
}

\begin{document}

\newcolumntype{x}[1]{>{\centering\arraybackslash}p{#1pt}}
\newcolumntype{y}[1]{>{\raggedright\arraybackslash}p{#1pt}}
\newcolumntype{z}[1]{>{\raggedleft\arraybackslash}p{#1pt}}
\maketitle
\begin{abstract}
This work delves into the task of pose-free novel view synthesis from stereo pairs, a challenging and pioneering task in 3D vision. 
Our innovative framework, unlike any before, seamlessly integrates 2D correspondence matching, camera pose estimation, and NeRF rendering, fostering a synergistic enhancement of these tasks. We achieve this through designing an architecture that utilizes a shared representation, which serves as a foundation for enhanced 3D geometry understanding. Capitalizing on the inherent interplay between the tasks, our unified framework is trained end-to-end with the proposed training strategy to improve overall model accuracy. Through extensive evaluations across diverse indoor and outdoor scenes from two real-world datasets, we demonstrate that our approach achieves substantial improvement over previous methodologies, especially in scenarios characterized by extreme viewpoint changes and the absence of accurate camera poses. The project page and code will be made available at: \url{https://ku-cvlab.github.io/CoPoNeRF/}.

\end{abstract}    
\section{Introduction}



In real-world scenarios aimed at rendering novel views from unposed images, the initial step often involves employing an off-the-shelf camera pose estimation~\cite{sarlin2020superglue,truong2023pdc,rockwell20228,zhang2022relpose}. These estimated poses are then typically integrated with a pre-trained generalized NeRF model~\cite{yu2021pixelnerf,du2023learning} to facilitate view synthesis. However, this approach is not without its drawbacks. The primary limitation stems from the inherent disparities or misalignments that may arise when combining models dedicated to different tasks. This method risks potential inconsistencies, as it treats pose estimation and NeRF rendering as distinct, separate processes, potentially leading to suboptimal results in the synthesized views.
\begin{figure}
    \centering\includegraphics[width=0.99\linewidth]{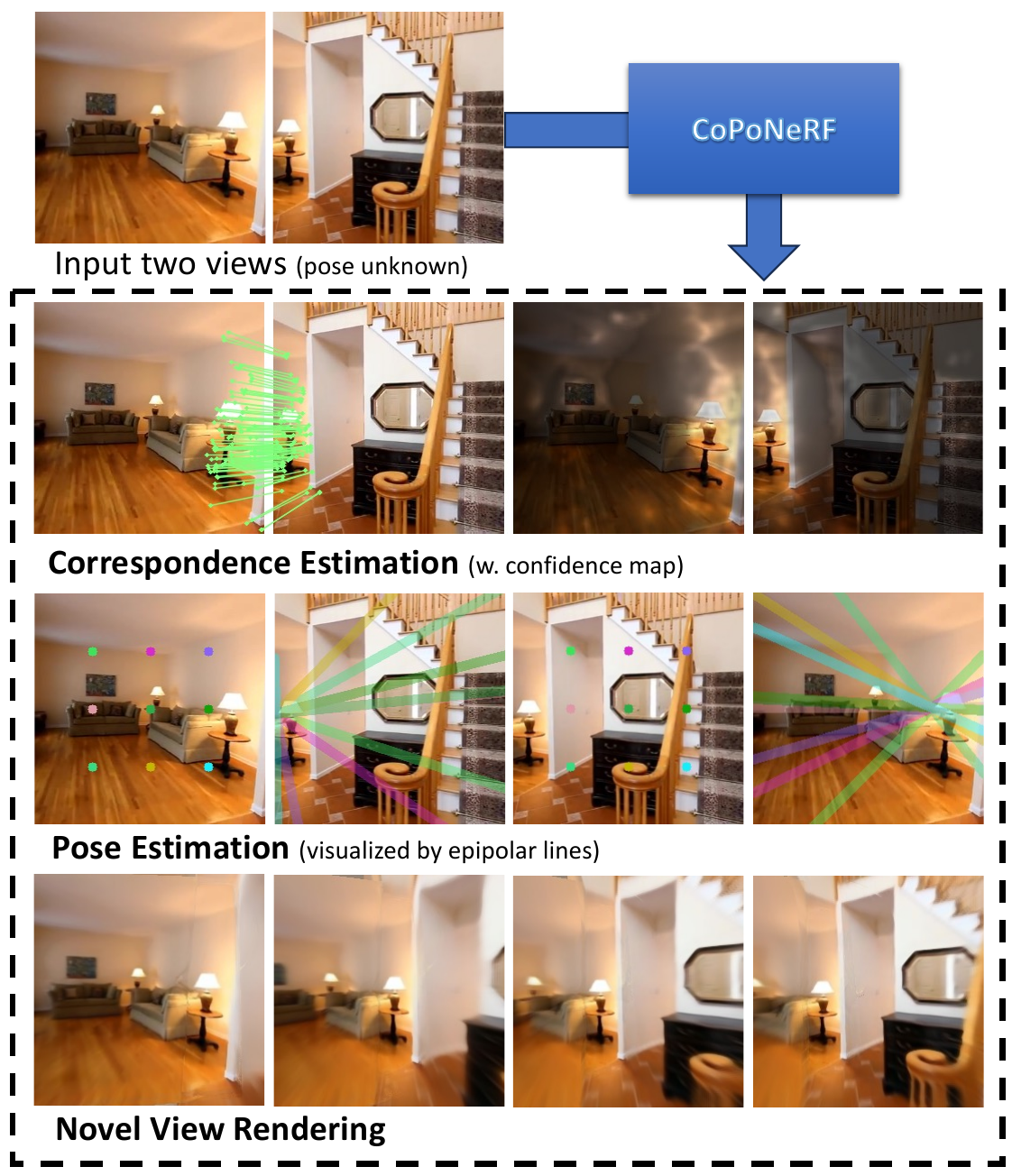}
    \vspace{-8pt}
    \caption{
    \textbf{Overview.} Given an unposed pair of images, possibly under extreme viewpoint changes and with minimal overlapping, our framework synergistically performs and effectively fosters mutual enhancement among three tasks -- 2D correspondence estimation, camera pose estimation, and NeRF rendering -- to enable high-quality novel view synthesis.   
    }
    \label{fig1}\vspace{-15pt}
\end{figure}

Recent developments in alternative approaches have trended towards the integration of pose estimation with NeRF rendering, thus leading to the advent of pose-free, generalized NeRF approaches~\cite{smith2023flowcam,chen2023dbarf}. This has been primarily realized through the meticulous assembly of developed modules in a multi-task framework. For example, \cite{smith2023flowcam} exploited correspondence information sourced from a well-established RAFT optical flow model~\cite{teed2020raft} and depth information from a single-view generalizable NeRF model~\cite{yu2021pixelnerf} for pose estimation. \cite{chen2023dbarf} combined a generalized NeRF module with a RAFT-like recurrent GRU module, responsible for camera pose and depth estimation, and implemented a three-stage training scheme for these two modules. Despite the promising results shown by these seminal approaches, the intrinsic complementarity of the three key tasks, correspondence estimation, pose estimation, and NeRF rendering, are not fully recognized and utilized. This resulted in solutions that were suboptimal, particularly in scenarios characterized by extreme viewpoint changes or minimal overlapping regions.

Acknowledging the shared core objectives between the three tasks, which is the precise interpretation and reconstruction of three-dimensional geometry from two dimensional image data, we emphasize the critical importance of cultivating a \textit{shared representation} among them. To this end, we propose a unified framework, namely CoPoNeRF, designed to estimate three distinct outputs, correspondence, camera pose, and NeRF rendering from this common representation. By adopting joint training, we maximize the synergy between these components, ensuring that each task not only contributes to but also benefits from this shared medium. This integrated approach effectively pushes the boundaries beyond what is achievable when treating each task as an independent and disjoint problem.

We evaluate the effectiveness of our framework using large-scale real-world indoor and outdoor datasets~\cite{liu2021infinite,zhou2018stereo}. Our results demonstrate that this framework successfully synthesizes high-quality novel views while simultaneously achieving precise relative camera pose estimation. We also provide extensive ablation studies to validate our choices. 
\textbf{Our contributions} are summarized as follows:
\begin{itemize}
 
  \item We tackle the challenging task of pose-free generalizable novel view synthesis, addressing the minimal view overlap scenarios that are not considered by prior methods. This aspect of our approach illustrates its applicability in handling complex, real-world conditions.

\item We propose a unified framework that enhances the processes of pose estimation, correspondence estimation, and NeRF rendering. This framework is designed to exploit the interdependencies of these components with a shared representation learning.

\item Leveraging the advanced representations learned by our framework, we achieve state-of-the-art performance not only in pose-free scenarios but also in generalized novel view synthesis with poses. 
\end{itemize}

%




\section{Related Work}
\paragraph{Generalized Neural Radiance Fields.}
Classical NeRF methodologies rely on numerous multi-view image datasets~\cite{mildenhall2021nerf, barron2021mip}, while recent efforts aim to learn reliable radiance fields from sparse imagery with a single feed-forward pass~\cite{yu2021pixelnerf,wang2021ibrnet,chen2021mvsnerf,johari2022geonerf,chen2023explicit,du2023learning}. These, however, depend heavily on precise camera poses and significant view overlap. To lessen this dependency, various frameworks optimize NeRF by integrating geometry and camera pose refinement, offering a degree of pose flexibility~\cite{jeong2021self,lin2021barf,zhu2022nice,bian2023nope,truong2023sparf}. \begin{figure*}
    \centering
    \includegraphics[width=1\linewidth]{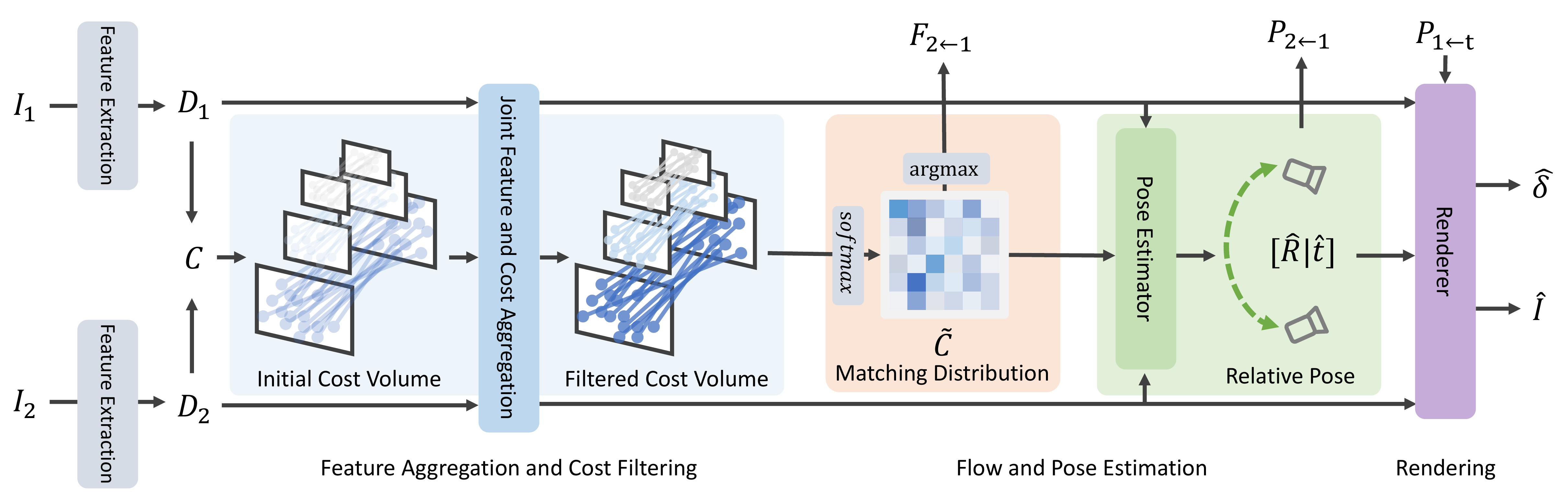}\hfill\\
    \vspace{-5pt}
    \caption{\textbf{Overall architecture of the proposed method.} For a pair of images, we extract multi-level feature maps and construct 4D correlation maps at each level, encoding pixel pair similarities. These maps are refined for flow and pose estimation, and the renderer then uses the estimated pose and refined feature maps for color and depth computation. }
    \label{fig:overall}\vspace{-10pt}
\end{figure*}

We focus on generalized frameworks for pose-free view synthesis; DBARF~\cite{chen2023dbarf}, for example, proposes a pose-agnostic solution by combining camera pose estimation with novel view synthesis. However, the network is trained in a staged manner with a local cost volume to encode multi-view information, struggling with minimal overlapping pairs and failing to fully harness the potential synergy between pose estimation and NeRF.  FlowCAM~\cite{smith2023flowcam}, on the other hand, leverages a weighted Procrustes analysis~\cite{choy2020deep} and an established optical flow network for point correspondences~\cite{teed2020raft}. Despite its attempt to formulate a multi-task framework, the reliance on the flow model inevitably risks failures in both view synthesis and pose estimation, especially for images with extreme viewpoint changes. \vspace{-12pt}  

\paragraph{Establishing Correspondences.}
Correspondence estimation, pivotal for various applications such as SLAM~\cite{durrant2006simultaneous}, SfM~\cite{schonberger2016structure}, and camera pose estimation~\cite{nister2004efficient}, traditionally entails a sequence involving keypoint detection~\cite{lowe2004distinctive,bay2006surf,rublee2011orb,detone2018superpoint}, feature description~\cite{yi2016lift,dusmanu2019d2,revaud2019r2d2}, tentative matching, and outlier filtering~\cite{yi2018learning,wei2023generalized,brachmann2017dsac,barath2020magsac++,barath2019magsac}. While outlier filtering stage also holds significant importance in relative pose estimation~\cite{barroso2023two}, the intrinsic quality of feature descriptors and the validity of matching scores markedly influence the pose prediction outcomes~\cite{cai2021extreme,rockwell20228}.
In this research, we harness the power of meticulously established correspondences, leveraging them to bolster both pose estimation and neural rendering processes, optimizing the overall task efficacy. \vspace{-12pt}

\paragraph{Camera Pose Estimation.}
Classic camera pose estimation methods primarily utilize hand-crafted algorithms to solve pose estimations using a set of correspondences, focusing on improving descriptor quality, cost volume, or outlier filtering to enhance correspondence quality~\cite{nister2004efficient,longuet1981computer,hartley1997defense}. More recent works have shifted towards learning direct mappings from images to poses. Notable advancements include the use of CNN-based networks to solve pose regression, such as the work by~\cite{melekhov2017relative} and subsequent developments~\cite{laskar2017camera,en2018rpnet}. Our work aligns more closely with methodologies tackling wide-baseline image pairs as inputs, an aspect relatively lesser explored. Some examples include leveraging a 4D correlation map for relative pose regression~\cite{cai2021extreme}, predicting discrete camera position distributions~\cite{chen2021wide}, and modifying the ViT~\cite{dosovitskiy2020image} to emulate the 8-point algorithm~\cite{rockwell20228}. Unique in approach, our method pioneers addressing the wide-baseline setting in generalized pose-free novel view synthesis tasks.
\section{Unified Framework for Generalized Pose-Free Novel View Synthesis}

\subsection{Problem Formulation}\label{sec:3.1}
Assuming an unposed pair of images $I_1, I_2 \in \mathbb{R}^{H \times W \times 3}$ taken from different viewpoints as the input, our goal is to synthesize an image $\hat{I}_t$ from a novel view. In this work, we assume camera intrinsics are given, as it is generally available from modern devices~\cite{arnold2022map}. Different from classical generalized NeRF tasks~\cite{wang2021ibrnet,chen2021mvsnerf,yu2021pixelnerf}, our task is additionally challenged by the absence of camera pose between the input images. To this end, we estimate the relative camera pose between $I_1, I_2$ as ${P}_{2\leftarrow 1} \in \mathbb{R}^{4\times 4},$ consisting of rotation $R \in \mathbb{R}^{3\times 3}$ and translation $T \in \mathbb{R}^{3\times 1}$, and deduce $P_{2\leftarrow t} = P_{2\leftarrow 1} P_{1\leftarrow t}$ with  $P_{1\leftarrow t}$ as the desired rendering viewpoint, which are then used in conjunction with the extracted feature maps to compute the pixel color at the novel view by the renderer.

\subsection{Cost Volume Construction }\label{sec:3.2}
The first stage of the pipeline is feature extraction, which will be shared across all three tasks. Because our method must be robust for scale differences and extreme viewpoint changes, we use multi-level feature maps to capture both geometric and semantic cues from different levels of features. Given a pair of images $I_1$ and $I_2$, we first extract $l$-levels of deep local features $D_1^l, D_2^l \in \mathbb{R}^{h^l \times w^l \times c^l}$ with ResNet~\cite{he2016deep}. Subsequent to feature extraction, the extracted features undergo cost volume construction. 

Unlike the previous methods that only consider local receptive fields within their cost volumes~\cite{yao2018mvsnet,chen2021mvsnerf,johari2022geonerf,chen2023dbarf}, we consider \emph{all pairs of similarities between features} to handle both small and large displacements~\cite{cho2021cats,cho2022cats++,hong2021deep,hong2022cost,hong2022neural}. Specifically, we compute and store the pairwise cosine similarity between features, obtaining a 4D cost volume  $\{C^l\}_{l=1}^L \in \mathbb{R}^{h^l\times w^l \times h^l \times w^l}$, where $L$ is the number of levels.  

\subsection{Feature Aggregation and Cost Filtering}\label{sec:3.3}
\paragraph{Joint Feature and Cost Aggregation.}
Recent progress in image correspondence has demonstrated the value of self- and cross-attention mechanisms in capturing global context within images  and enhancing inter-image feature extraction, vital for understanding multi-view geometry \cite{xu2022gmflow, du2023learning, sun2021loftr}. Studies have also emphasized the importance of cost aggregation for reducing noise in cost volumes and embedding geometric priors \cite{chen2023explicit, cho2023cat, huang2022flowformer}.


Building upon these developments, we introduce a self-attention-based aggregation block that processes the feature maps and cost volume, \ie, $D_1$, $D_2$, and $C$ (level indicator $l$ omitted for brevity). Specifically, two augmented cost volumes are first constructed by feature and cost volume concatenation: $[C, D_1]$ and $[C^T, D_2]\in \mathbb{R}^{h\times w \times (hw+c)}$. Then, treating each 2D location in the augmented cost volume as a token, our aggregation block performs self-attention operation $\phi$ using feature maps and cost volumes as \textit{values}. The resulting cost volumes are obtained as $\phi ([C, D_1]) + \phi ([C^T, D_2])^T$ that ensures consistent matching scores.
\vspace{-12pt}

\paragraph{Leveraging Cost Volume as Matching Distribution. }

Our method leverages enhanced feature maps and a refined cost volume from the aggregation block to inter-condition the feature maps.
Unlike the standard practice of using a cross-attention map from two feature maps \cite{sun2021loftr, xu2022gmflow, chen2023explicit}, we introduce a simple and more effective adaptation by employing the refined cost volume from the aggregation block, rather than computing a separate cross-attention map. 
Specifically, we apply softmax to this volume to create a cross-attention map, which then guides the alignment of feature maps with matching probabilities. This layer is integrated with the aggregation block in an interleaved manner, crucial for refining and assimilating multi-view information. More details can be found in the \emph{supp. material}.

The final cost volume, $\frac{1}{L}\sum_l^L \tilde{C}^l$, calculated from each $l$-th level, is then used for relative pose and flow estimation. This cost volume plays a pivotal role in consolidating multi-level feature correspondences, directly impacting the accuracy of our pose and flow estimations. 


\subsection{Flow and Relative Pose Estimation}\label{sec:3.4}
Making use of the cost volume from previous steps, we define a dense flow field, $F(i)$ that warps all pixels $i$ in image $I_1$ towards $I_2$. We also estimate relative camera pose $P_{2\leftarrow 1}$ from this cost volume as it sufficiently embodies confidence scores and spatial correspondences~\cite{cai2021extreme,rockwell20228}. To estimate the dense flow map $F_{2 \leftarrow 1}$, we can simply apply argmax to find the highest scoring correspondences. While this may be sufficient for image pairs with large overlapping regions, we account for the potential occlusions by computing a confidence map. Specifically, following~\cite{meister2018unflow}, we obtain a cyclic consistency map $M_{2\leftarrow 1}(i)$ using the reverse field $F_{1 \leftarrow 2}$ as an additional input, and check if the following condition is met for consistency: $||F_{2\leftarrow 1}(i) + F_{1 \leftarrow 2}(i + F_{2\leftarrow 1}(i))||_2 < \tau$, where $||\cdot ||_2$ denotes frobenius norm and $\tau$ is a threshold hyperparameter. The reverse cyclic consistency map  $M_{1\leftarrow 2}$ is computed with similar procedure. 

\begin{figure}[t]
\begin{center}
\begin{subfigure}{.091\textwidth}
  \centering
  \includegraphics[width=1.0\linewidth]{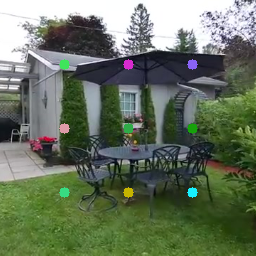}
\end{subfigure}
\begin{subfigure}{.091\textwidth}
  \centering
  \includegraphics[width=1.0\linewidth]{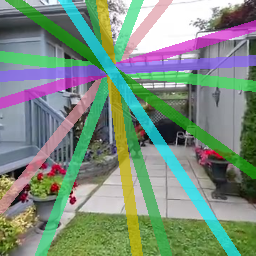}
\end{subfigure}
\begin{subfigure}{.091\textwidth}
  \centering
  \includegraphics[width=1.0\linewidth]{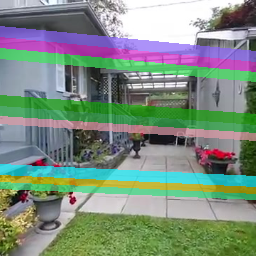}
\end{subfigure}
\begin{subfigure}{.091\textwidth}
  \centering
  \includegraphics[width=1.0\linewidth]{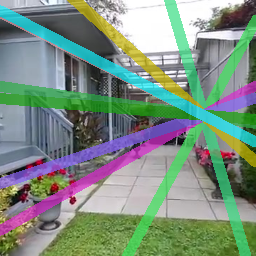 }
\end{subfigure}
\begin{subfigure}{.091\textwidth}
  \centering
  \includegraphics[width=1.0\linewidth]{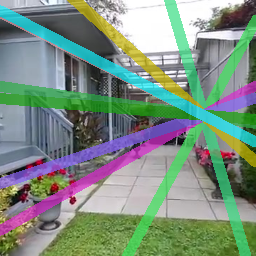}
\end{subfigure}\\\vspace{1.5pt}
\begin{subfigure}{.091\textwidth}
  \centering
  \includegraphics[width=1.0\linewidth]{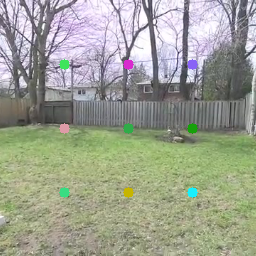}\caption{$I_1$}
\end{subfigure}
\begin{subfigure}{.091\textwidth}
  \centering
  \includegraphics[width=1.0\linewidth]{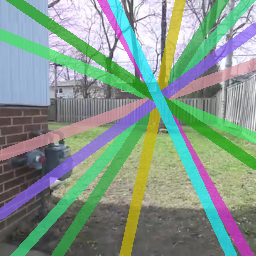}\caption{DBARF}
\end{subfigure}
\begin{subfigure}{.091\textwidth}
  \centering
  \includegraphics[width=1.0\linewidth]{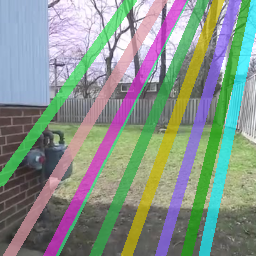}\caption{FlowCAM}
\end{subfigure}
\begin{subfigure}{.091\textwidth}
  \centering
  \includegraphics[width=1.0\linewidth]{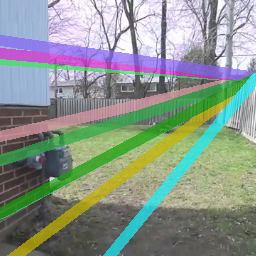 }\caption{Ours}
\end{subfigure}
\begin{subfigure}{.091\textwidth}
  \centering
  \includegraphics[width=1.0\linewidth]{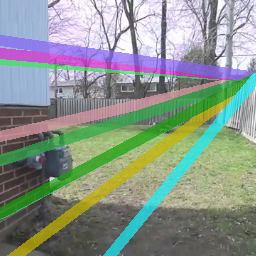}\caption{GT}
\end{subfigure}
\end{center}
\vspace{-15pt}
\caption{\textbf{Visualization of epipolar lines.} We use the relative camera pose to draw epipolar lines based on the points in (a). Our predictions can well follow the ground truth even under large viewpoint changes.  }
\label{fig:epipolar}\vspace{-10pt}
\end{figure}


To estimate camera parameters, we use the knowledge taken from previous study~\cite{rockwell20228} and adopt an essential matrix module to output rotation $R$ and translation $t$. The essential matrix module is a mapping module that exploits each transformer token from images to a feature that is used to predict $R$ and $t$. This module contains positional encoding, bilinear attention, and dual softmax over attention map $A$. Following the design in Sec.~\ref{sec:3.3}, we make a modification to replace $A$ with our cost volume, since it acts as a key for emulating 8-point algorithm, such that the better spatial correspondences encoded in $A$ can aid more accurate camera pose estimation. Subsequent to the essential matrix module, we finally regress 6D rotation representations~\cite{Zhou_2019_CVPR} and 3 translation parameters with scales using MLPs. 

\subsection{Attention-based Renderer}\label{sec:3.5}

Within our method, the rendering module is tasked with synthesizing novel views, guided by the estimated camera poses and a pair of aggregated features from previous steps. Borrowing from recent advancements, we adopt a strategy of sampling pixel-aligned features along the epipolar lines of each image and augment the features by a corresponding feature in the other image,  as suggested by Du et al.~\cite{du2023learning}. This technique also enables us to ascertain depth $\delta$ by triangulating these features, thereby streamlining the typically resource-intensive 3D sampling process. Given the set of sampled features from epipolar lines, we adopt an attention-based rendering procedure to compute the pixel color, as done similarly in previous methods~\cite{johari2022geonerf,du2023learning,suhail2022generalizable}.

\begin{figure}[t]
\begin{center}
\begin{subfigure}{.48\textwidth}
  \centering
  \includegraphics[width=1.0\linewidth]{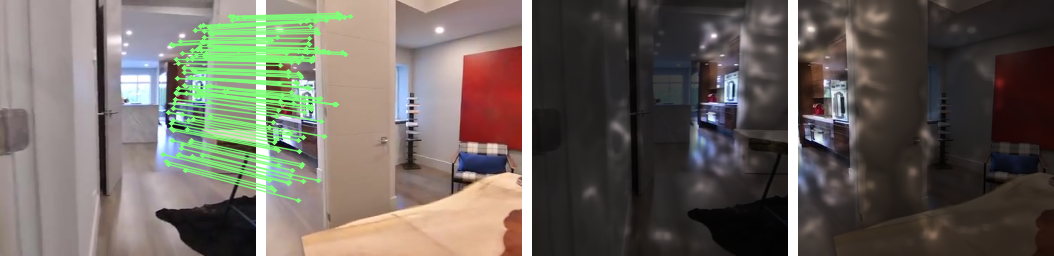}
\end{subfigure}\\\vspace{1.5pt}
\begin{subfigure}{.48\textwidth}
  \centering
  \includegraphics[width=1.0\linewidth]{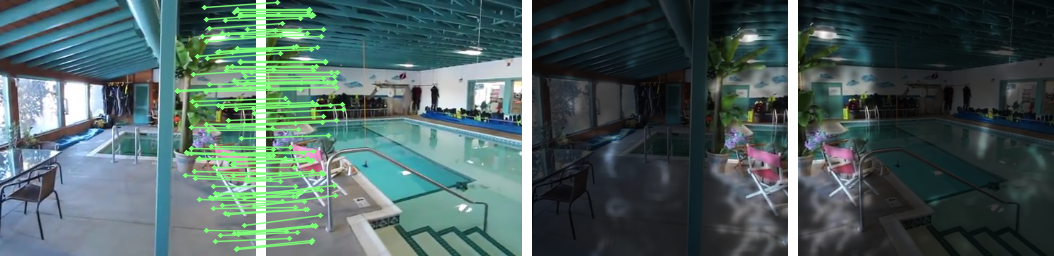}
\end{subfigure}
\end{center}
\vspace{-15pt}
\caption{\textbf{Visualization of correspondences and confidence. } We show top 100 confident matches between input images and the covisible regions are highlighted based on confidence scores.     }
\label{fig:epipolar}\vspace{-9pt}
\end{figure}

\subsection{Training Objectives}\label{sec:3.6}
The outputs of our models are colors, depths, an estimated relative camera pose, and a dense flow map. Our model is trained with an objective function consisting of four losses: image reconstruction loss $\mathcal{L}_\text{img}$, matching loss $\mathcal{L}_\text{match}$, camera pose loss $\mathcal{L}_\text{pose}$, and the triplet consistency loss $\mathcal{L}_\text{tri}$. For rendering, we use the photometric loss between the rendered color and the target color defined as $L1$ loss. \vspace{-12pt}
 
\paragraph{Matching Loss.} To provide training signals for correspondence estimation, we adopt self-supervised SSIM loss as a valuable alternative since obtaining ground-truth correspondences between image pairs is often challenging and impractical, since the depth information is required. The SSIM ~\cite{wang2004image} loss computes the structural similarity between the warped image and the target image, offering a data-driven approach to assess the quality of image registration without relying on explicit depth measurements or ground-truth correspondences. The matching loss is defined as: 
\begin{equation}
\begin{split}
    \mathcal{L}_\text{match} =
    \sum _i M_{1\leftarrow 2}(i)(1-\text{SSIM}(F_{1\leftarrow 2}(I_2(i)), I_{1}(i))) \\
    + M_{2\leftarrow 1}(i)(1-\text{SSIM}(F_{2\leftarrow 1}(I_1(i)), I_{2}(i))),
\end{split}
\end{equation}
where $\phi(\cdot, \cdot)$ yields the SSIM score between two comparative inputs.


\paragraph{Pose Loss. }
Although we only take a pair of unposed images as input for the inference phase, for the training phase, we incorporate readily available and ubiquitous camera poses, thanks to the extensive availability of video data and the deployment of conventional pose estimation methodologies prevalent in the field, including SfM and SLAM. Our pose loss is a combination of geodesic loss~\cite{salehi2018real} for rotation and $L_2$ distance loss for translation\footnote{Although two-view geometry inherently lacks the capability to discern translation scales, we let the model learn to align all predictions to true scales  via recognition, as done in \cite{rockwell20228}.}. Specifically, they are combined with addition and defined as:
\begin{equation}
\begin{split}
 &\mathcal{L}_{\text{rot}} =\arccos\left(\frac{\text{trace}(\hat{R}^T R) - 1}{2}\right)\\
 &\mathcal{L}_{\text{trans}}(\hat{t}, t) = \|\hat{t} - t\|^2_2,\\
\end{split}
\end{equation} 
where $\hat{R}$ and $\hat{t}$ indicates the estimated rotation and translation.

Empirical results indicate that including gradient feedback from other losses alongside the pose loss contributes to unstable training, typically leading to suboptimal model performance.  Aligning with the literature \cite{jeong2021self,lin2021barf,chen2023dbarf}, our findings also accentuate that the expansive search space and the intrinsic complexities associated with pose optimization increase the difficulty of the learning process.

To address this challenge, we  directly incorporate ground-truth pose data into key modules, like rendering or feature projection, during training. This approach restricts gradient flow to the pose loss, proving highly effective in our experiments. Conceptually, this resembles the teacher forcing strategy in RNNs~\cite{williams1989learning}, where ground truth, rather than previous network outputs, guides training. This method encourages network parameters are optimized in direct alignment with pose estimation objectives, similar to using ground truth inputs for more direct supervision in teacher forcing.\vspace{-15pt}

\

\begin{figure*}
    \centering
    \includegraphics[width=1\linewidth]{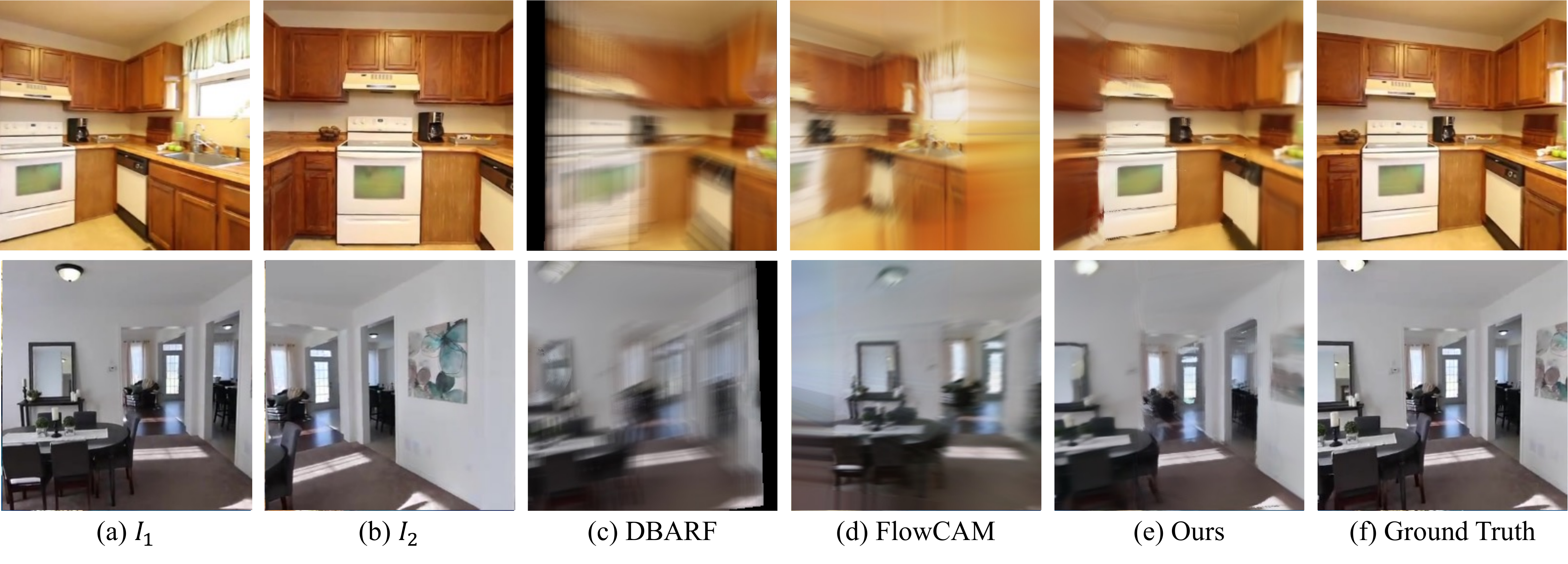}\hfill\\
    \vspace{-10pt}
    \caption{\textbf{Qualitative comparison on RealEstate10K.} }
    \label{fig:realestate}\vspace{-10pt}
\end{figure*}

\paragraph{Triplet Consistency Loss. }Finally, we propose a triplet consistency loss $\mathcal{L}_\text{tri}$ seamlessly incorporating all the outputs of our model, flow, pose, and depth. Our loss extends the cycle consistency~\cite{zhu2017unpaired} loss, which has been prevalent in the field of computer vision, by incorporating the three outputs from our network. Specifically, at each iteration, given a set of randomly selected coordinates in the target frame that are projected to $I_1$ and $I_2$ using depth $\delta$, camera pose $P$, and camera intrinsic $K$, we define the p: $i'_1 = K_1 P_{1\leftarrow t} \delta(i) K_1^{-1} i$ and $i'_2$ is defined similarly.  Using the set of projected points $i'_1$, we employ $F_{2\leftarrow 1}$ to warp them towards $I_2$ to obtain warped coordinates $\hat{i}'_2$. We finally apply Huber loss function~\cite{hastie2009elements} $\psi$ such that $M(i'_1)\psi (\hat{i}'_2, {i}'_2)$, where $M(\cdot)$ eliminates the unconfident matches. This loss effectively measures the consistency between the estimated depth and optical flow across the views. If the estimated depth and flow are accurate, the projected and warped points should coincide, resulting in a small loss value. 

In summary, our total training loss is  defined as
\begin{equation}
    \mathcal{L}_\text{final} = \mathcal{L}_\text{img} + \mathcal{L}_\text{match} + \mathcal{L}_\text{pose} + \lambda_{\text{tri}}\mathcal{L}_\text{tri},
\end{equation}
where $\lambda_{\text{tri}}$ is a scaling factor.
\section{Experiments}

\subsection{Implementation Details}
Our encoder uses ResNet-34, taking 256$\times$256 image as an input, and extracts 3 levels of feature maps with a spatial resolution of 16, 32, and 64. We set $\lambda_{\text{tri}} = 0.01$. Our network is implemented using PyTorch~\cite{paszke2017automatic} and trained with the AdamW~\citep{loshchilov2017decoupled} optimizer. We set the base learning rate as 2e$-4$ and use an effective batch size of 64. The network is trained for 50K iterations, taking approximately 2 days. We exponentially decay the learning rate with $\gamma = 0.95$ after every epoch. We train and evaluate all other baselines on the same datasets for fair comparisons. We provide training and evaluation details of ours and our competitors in the \emph{supp. material} for completeness. 
\subsection{Experimental Settings}

\paragraph{Datasets.}
We train and evaluate our method on RealEstate10K~\cite{zhou2018stereo}, a large-scale dataset of both indoor and outdoor scenes, and ACID~\cite{liu2021infinite}, a large-scale dataset of outdoor coastline scenes. For RealEstate10K, we employ a subset of the complete dataset, resulting in a training set comprising 21,618 scenes and a testing set consisting of 7,200 scenes, while for ACID, we use 10,935 scenes for training and 1,893 scenes for testing.  \vspace{-12pt}
\paragraph{Tasks and Baselines.}
We assess the performance of our method on two tasks: novel-view synthesis and relative pose estimation. The latter task serves a dual purpose as it also assesses the quality of our correspondences, a methodological approach that aligns with prior image matching studies. We first compare with established generalized NeRF variants, specifically PixelNeRF~\cite{yu2021pixelnerf} and~\cite{du2023learning}, highlighting the complexities of wide-baseline inputs. Subsequently, we compare with existing pose-free generalized NeRF methods, namely DBARF~\cite{chen2023dbarf} and FlowCAM~\cite{smith2023flowcam}. For relative pose estimation, the evaluation includes several comparisons:  matching methods~\cite{detone2018superpoint,sarlin2020superglue,truong2023pdc} followed by 5-point algorithm~\cite{nister2004efficient} and RANSAC~\cite{fischler1981random} and end-to-end pose estimation frameworks~\cite{zhang2022relpose, rockwell20228}. Finally, we conduct a comparative analysis with pose-free generalized NeRF methods~\cite{chen2023dbarf,smith2023flowcam}.\vspace{-12pt}

\paragraph{Evaluation Metrics.}
We use the standard image quality metrics (PSNR, SSIM, LPIPS, MSE) for novel view synthesis evaluation. For relative pose estimation, we use geodesic rotation error and angular difference for translation\footnote{Since translation scale is theoretically indeterminable in two-view camera pose estimation, evaluating it could potentially lead to inconclusive or erroneous interpretations (see the \emph{supp. material} for more results).} as done in classical methods \cite{nister2004efficient,melekhov2017relative}. Our statistical analysis includes average, median, and standard deviation of errors, with the median offering robustness against outliers and standard deviation indicating error variability.\vspace{-12pt}

\paragraph{Evaluation Protocol.}
As our approach is the first work to this extremely challenging task, we introduce a new evaluation protocol. By default, we assume $I_1, I_2$ and $P_{1\leftarrow t}$ are provided to the model, while \emph{$I_t$ is used solely for computing the metrics}\footnote{Existing pose-free generalized NeRF methods use target frames for additional geometric cues during evaluation \cite{chen2023dbarf,smith2023flowcam}. For practicality, we assume target frames are \emph{only available for metric calculation}, not for method operation, applying this uniformly across all methods. This aligns with real-world scenarios where target views are not accessible.}. First, when presented with a video sequence of a scene, we employ a frame-skipping strategy. The value of $N$ frames skipped between each frame is dynamically determined based on the total number of frames. For sequences with fewer than 100 frames, $N$ is calculated as one-third of the total frame count; otherwise, we set $N=50$. This gives us three images $I_1, I_t, I_2$, which are taken from $N = 0, 50, 100,$ respectively. Next, for evaluation, while a common practice in relative pose estimation tasks is to leverage rotation variance~\cite{cai2021extreme,rockwell20228}, this approach disregards translations, often leading to controversial classification of the images into one of the distributions in some cases, \textit{e.g.,} image pairs with zoomed-in and out cameras. We thus partition the test set into three subsets named \textit{Small}, \textit{Medium}, and \textit{Large} to denote the \emph{degree of overlap in the scenes}. 
With such a splitting scheme, for the RealEstate10K dataset, we obtain subsets containing 3593, 1264, and 2343 scenes respectively, whereas for the ACID dataset, they encompass 559, 429, and 905 scenes each. We show the visualization of the splits and their distributions in the \emph{supp. material}.

\begin{table*}[]
\scalebox{0.64}{
    \centering
    \begin{tabular}{l|c|l|ccc|ccc|ccc|ccc}
    \toprule
         \multirow{3}{*}{Overlap}&\multirow{3}{*}{Task}&\multirow{3}{*}{Method} & \multicolumn{6}{c|}{RealEstate-10K} &\multicolumn{6}{c}{ACID} \\\cline{4-15}
         && &\multicolumn{3}{c}{Rotation} &\multicolumn{3}{c|}{Translation} &\multicolumn{3}{c}{Rotation} &\multicolumn{3}{c}{Translation}\\
        & & &Avg(\degree $\downarrow$)& Med(\degree $\downarrow$) &STD(\degree $\downarrow$) 
         &Avg(\degree $\downarrow$)& Med(\degree $\downarrow$) &STD(\degree $\downarrow$)& Avg(\degree $\downarrow$)&Med(\degree $\downarrow$) &STD(\degree $\downarrow$) 
         &Avg(\degree  $\downarrow$)& Med(\degree  $\downarrow$) &STD(\degree  $\downarrow$)\\\midrule 
         \multirow{7}{*}{Small} &{{COLMAP}}&\color{gray}{SP+SG}~\cite{detone2018superpoint,sarlin2020superglue,fischler1981random}&\color{gray}{9.793}&\color{gray}{2.270}&\color{gray}{22.084}&\color{gray}{12.549}&\color{gray}{4.638}&\color{gray}{23.048}&\color{gray}{10.920}&\color{gray}{2.797}&\color{gray}{22.761}&\color{gray}{22.214}&\color{gray}{7.526}&\color{gray}{33.719}\\
          &(Matching)&\color{gray}{PDC-Net+}~\cite{fischler1981random,truong2023pdc} &\color{gray}{3.460}&\color{gray}{1.128}&\color{gray}{7.717}&\color{gray}{6.913}&\color{gray}{2.752}&\color{gray}{15.558}&\color{gray}{2.520}&\color{gray}{0.579}&\color{gray}{6.372}&\color{gray}{15.664}&\color{gray}{4.215}&\color{gray}{29.640}\\\cline{2-15}
         &\multirow{2}{*}{Pose Estimation}&Rockwell \textit{et al.}~\cite{rockwell20228} &12.604&6.860&14.502&91.455&91.499&56.872&8.466&3.151&13.380&88.421&88.958&36.212\\
         &&RelPose~\cite{zhang2022relpose}&12.102&4.803&21.686&-&-&-&10.081&4.753&13.343&-&-&-\\\cline{2-15}
         &\multirow{3}{*}{Pose-Free NeRF}&DBARF~\cite{chen2023dbarf}&17.520&13.218&15.946&126.282&140.358&43.691&8.721&3.205&12.916&95.149&99.490&47.576\\
         &&FlowCAM~\cite{smith2023flowcam}& 11.883 & 6.778 & 15.676 & 87.119 & 58.245 & 26.895& 8.663 & 6.675 & 7.930 & 92.130 & 85.846 & 40.821\\\cline{3-15}
         &&Ours&\textbf{5.471}&\textbf{2.551}&\textbf{11.733}&\textbf{11.862}&\textbf{5.344}&\textbf{21.080}&\textbf{3.548}&\textbf{1.129}&\textbf{8.619}&\textbf{23.689}&\textbf{11.289}&\textbf{30.391}
         \\\midrule
         \multirow{7}{*}{Medium} &{{COLMAP}}&\color{gray}{SP+SG}~\cite{detone2018superpoint,sarlin2020superglue,fischler1981random}&\color{gray}{1.789}&\color{gray}{0.969}&\color{gray}{3.502}&\color{gray}{9.295}&\color{gray}{3.279}&\color{gray}{20.456}&\color{gray}{3.275}&\color{gray}{1.306}&\color{gray}{6.474}&\color{gray}{16.455}&\color{gray}{5.426}&\color{gray}{29.035}\\
         &(Matching)  &\color{gray}{PDC-Net+}~\cite{fischler1981random,truong2023pdc} &\color{gray}{1.038}&\color{gray}{0.607}&\color{gray}{1.841}&\color{gray}{6.667}&\color{gray}{2.262}&\color{gray}{18.247}&\color{gray}{2.378}&\color{gray}{0.688}&\color{gray}{5.841}&\color{gray}{14.940}&\color{gray}{4.301}&\color{gray}{27.379}\\\cline{2-15}
         &\multirow{2}{*}{Pose Estimation}&Rockwell \textit{et al.}~\cite{rockwell20228} &12.168&6.552&14.385&82.478&82.920&55.094&4.325&1.564&6.177&90.555&90.799&51.469\\
         &&RelPose~\cite{zhang2022relpose}&4.942&3.476&6.206&-&-&-&5.801&2.803&6.574&-&-&-\\\cline{2-15}
         &\multirow{3}{*}{Pose-Free NeRF}&DBARF~\cite{chen2023dbarf}&7.254&4.379&7.009&79.402&75.408&54.485&4.424&1.685&6.164&77.324&77.291&49.735\\
         &&FlowCAM~\cite{smith2023flowcam}& 4.154 & 3.346 & 3.466 & 42.287 & 41.594 & 24.862& 8.778 & 6.589 & 7.489 & 95.444 & 87.308 & 43.198\\\cline{3-15}
         &&Ours&\textbf{2.183}&\textbf{1.485}&\textbf{2.419}&\textbf{10.187}&\textbf{5.749}&\textbf{15.801}&\textbf{2.573}&\textbf{1.169}&\textbf{3.741}&\textbf{21.401}&\textbf{10.656}&\textbf{28.243}\\\bottomrule
          \multirow{7}{*}{Large} &{{COLMAP}}&\color{gray}{SP+SG}~\cite{detone2018superpoint,sarlin2020superglue,fischler1981random}&\color{gray}{1.416}&\color{gray}{0.847}&\color{gray}{1.984}&\color{gray}{21.415}&\color{gray}{7.190}&\color{gray}{34.044}&\color{gray}{1.851}&\color{gray}{0.745}&\color{gray}{3.346}&\color{gray}{22.018}&\color{gray}{7.309}&\color{gray}{33.775}\\
           &(Matching)&\color{gray}{PDC-Net+}~\cite{fischler1981random,truong2023pdc} &\color{gray}{0.981}&\color{gray}{0.533}&\color{gray}{1.938}&\color{gray}{16.567}&\color{gray}{5.447}&\color{gray}{29.883}&\color{gray}{1.953}&\color{gray}{0.636}&\color{gray}{4.133}&\color{gray}{18.447}&\color{gray}{4.357}&\color{gray}{35.564}\\\cline{2-15}
         &\multirow{2}{*}{Pose Estimation} &Rockwell \textit{et al.}~\cite{rockwell20228} &12.771&7.214&14.863&91.851&88.923&57.444&2.280&0.699&3.512&86.580&87.559&50.369\\
         &&RelPose~\cite{zhang2022relpose}&4.217&2.447&5.621&-&-&-&4.309&2.011&5.288&-&-&-\\\cline{2-15}
         &\multirow{3}{*}{Pose-Free NeRF}&DBARF~\cite{chen2023dbarf}&3.455&1.937&3.862&50.094&33.959&43.659&2.303 & 0.859 & 3.409 & 54.523 & 38.829 & 45.453\\
         &&FlowCAM~\cite{smith2023flowcam}& 2.349 & 1.524 & 2.641 & 34.472 & 27.791 & 31.615& 9.305 & 6.898 & 9.929 & 97.392 & 89.359 & 43.777\\\cline{3-15}
         &&Ours&\textbf{1.529}&\textbf{0.991}&\textbf{1.822}&\textbf{15.544}&\textbf{7.907}&\textbf{24.626}&\textbf{3.455}&\textbf{1.129}&\textbf{7.265}&\textbf{22.935}&\textbf{10.588}&\textbf{30.974}\\\bottomrule
         \multirow{7}{*}{\textit{Avg}} &\multirow{2}{*}{{COLMAP}}&\color{gray}{SP+SG}~\cite{detone2018superpoint,sarlin2020superglue,fischler1981random} &\color{gray}{5.605}&\color{gray}{1.301}&\color{gray}{16.129}&\color{gray}{14.887}&\color{gray}{5.058}&\color{gray}{27.238}&\color{gray}{4.819}&\color{gray}{1.203}&\color{gray}{13.473}&\color{gray}{20.802}&\color{gray}{6.878}&\color{gray}{32.834}\\
        &(Matching) &\color{gray}{PDC-Net+}~\cite{fischler1981random,truong2023pdc}&\color{gray}{2.189}&\color{gray}{0.751}&\color{gray}{5.678}&\color{gray}{10.100}&\color{gray}{3.243}&\color{gray}{22.317}&\color{gray}{2.315}&\color{gray}{0.619}&\color{gray}{5.655}&\color{gray}{16.461}&\color{gray}{4.292}&\color{gray}{31.391}\\\cline{2-15}
         &\multirow{2}{*}{Pose Estimation}&Rockwell \textit{et al.}~\cite{rockwell20228} &12.585&6.881&14.587&90.115&88.648&40.948&4.568&1.312&8.358&88.433&88.961&36.197\\
         &&RelPose~\cite{zhang2022relpose}&8.285&3.845&16.329&-&-&-&6.348&2.567&9.047&-&-&-\\\cline{2-15}
         &\multirow{3}{*}{Pose-Free NeRF}&DBARF~\cite{chen2023dbarf}&11.144&5.385&13.516&93.300&102.467&57.290&4.681 & 1.421 & 8.417 & 71.711 & 68.892 & 50.277\\
         &&FlowCAM~\cite{smith2023flowcam}&7.426 & 4.051 & 12.135 & 50.659 & 46.281 & 52.321& 9.001 & 6.749 & 8.864 & 95.405 & 88.133 & 42.849\\\cline{3-15}
         &&Ours&\textbf{3.610}&\textbf{1.759}&\textbf{8.617}&\textbf{12.766}&\textbf{7.534}&\textbf{15.510}&\textbf{3.283}&\textbf{1.134}&\textbf{7.093}&\textbf{22.809}&\textbf{14.502}&\textbf{21.572}\\\bottomrule
    \end{tabular}
    }\vspace{-6pt}
    \caption{\textbf{Pose estimation performance on RealEstate-10K and ACID.} Gray color indicates methods not directly comparable as they supervise correspondence with ground-truth depth; they are included for reference only. We also specify the targeted task for each method.  
    }\vspace{-9pt}
    \label{tab:2}
\end{table*}

\begin{table}[]
\centering
\scalebox{0.56}{
    \begin{tabular}{c|c|l|cccc|cccc}
    \toprule
         \multirow{2}{*}{\!\!\!Overlap\!\!\!}&GT&\multirow{2}{*}{\!\!Method} & \multicolumn{4}{c|}{RealEstate-10K} &\multicolumn{4}{c}{ACID} \\\cline{4-11}
        &\!Pose\!& & \!\!\!{\small PSNR$\uparrow$}\!\!\! &\!\!\!{\small LPIPS$\downarrow$}\!\!\! &\!\!\!{\small SSIM$\uparrow$}\!\!\!  & \!\!\!{\small MSE$\downarrow$}\!\!\! &\!\!\!{\small PSNR$\uparrow$}\!\!\!  &\!\!\!{\small LPIPS$\downarrow$}\!\!\! &\!\!\!{\small SSIM$\uparrow$}\!\!\!  & \!\!\!{\small MSE$\downarrow$}\!\!\!
         \\\midrule
          \multirow{5}{*}{\!\!Small\!\!}&\multirow{2}{*}{\!\!\cmark\!\!}&\!\!\color{gray}{PixelNeRF}~\cite{yu2021pixelnerf}\!\! &\color{gray}{13.126}&\color{gray}{0.639}&\color{gray}{0.466}&\color{gray}{0.058}&\color{gray}{16.996}&\color{gray}{0.528}&\color{gray}{0.487}&\color{gray}{0.030}\\
          &&\color{gray}{\!\!Du \textit{et al.}}~\cite{du2023learning} &\color{gray}{18.733}&\color{gray}{0.378}&\color{gray}{0.661}&\color{gray}{0.018}&\color{gray}{25.553}&\color{gray}{0.301}&\color{gray}{0.773}&\color{gray}{0.005}\\\cline{2-11}
          &\multirow{3}{*}{\!\!\xmark\!\!}&\!\!DBARF~\cite{chen2023dbarf} &13.453&0.563&0.522&0.045&14.306&0.503&0.541&0.037\\
          &&\!\!FlowCAM~\cite{smith2023flowcam}\!\!& 15.435 & 0.528 & 0.570 & 0.034& 20.153 & 0.475 & 0.594 & 0.016\\
       \cline{3-11}
          &&\!\!Ours &\textbf{17.153}&\textbf{0.459}&\textbf{0.577}&\textbf{0.025}&\textbf{22.322}&\textbf{0.358}&\textbf{0.649}&\textbf{0.010}\\\midrule
           \multirow{5}{*}{\!\!Medium\!\!} &\multirow{2}{*}{{\!\!\cmark\!\!}}&\!\!\color{gray}{PixelNeRF}~\cite{yu2021pixelnerf}\!\! &\color{gray}{13.999}&\color{gray}{0.582}&\color{gray}{0.462}&\color{gray}{0.042}&\color{gray}{17.228}&\color{gray}{0.534}&\color{gray}{0.501}&\color{gray}{0.029}\\
          &&\color{gray}{\!\!Du \textit{et al.}}~\cite{du2023learning} &\color{gray}{22.552}&\color{gray}{0.263}&\color{gray}{0.764}&\color{gray}{0.008}&\color{gray}{25.694}&\color{gray}{0.303}&\color{gray}{0.769}&\color{gray}{0.005}\\\cline{2-11}
          &\multirow{3}{*}{\!\!\xmark\!\!}&\!\!DBARF~\cite{chen2023dbarf} &15.201&0.487&0.560&0.030&14.253&0.457&0.538&0.038\\
          &&\!\!FlowCAM~\cite{smith2023flowcam}\!\!& 18.481 & 0.592 & 0.441 & 0.018& 20.158 & 0.476 & 0.585 & 0.015\\
         \cline{3-11}
          &&\!\!Ours &\textbf{19.965}&\textbf{0.343}&\textbf{0.645}&\textbf{0.013}&\textbf{22.407}&\textbf{0.352}&\textbf{0.648}&\textbf{0.009}\\\midrule
           \multirow{5}{*}{Large} &\multirow{2}{*}{{\!\!\cmark\!\!}}&\!\!\color{gray}{PixelNeRF}~\cite{yu2021pixelnerf}\!\! &\color{gray}{15.448}&\color{gray}{0.479}&\color{gray}{0.470}&\color{gray}{0.031}&\color{gray}{17.229}&\color{gray}{0.522}&\color{gray}{0.500}&\color{gray}{0.028}\\
          &&\color{gray}{\!\!Du \textit{et al.}}~\cite{du2023learning} &\color{gray}{26.199}&\color{gray}{0.182}&\color{gray}{0.836}&\color{gray}{0.004}&\color{gray}{25.338}&\color{gray}{0.307}&\color{gray}{0.763}&\color{gray}{0.005}\\\cline{2-11}
          &\multirow{3}{*}{\!\!\xmark\!\!}&\!\!DBARF~\cite{chen2023dbarf} &16.615&0.380&0.648&0.022&14.086&0.419&0.534&0.039\\
          &&\!\!FlowCAM~\cite{smith2023flowcam}\!\!&22.418 & 0.707 & 0.287 & 0.009& 20.073 & 0.478 & 0.580 & 0.016\\
        \cline{3-11}
          &&Ours &\textbf{22.542}&\textbf{0.250}&\textbf{0.724}&\textbf{0.008}&\textbf{22.529}&\textbf{0.351}&\textbf{0.649}&\textbf{0.009}\\\midrule
          \multirow{5}{*}{\textit{Avg}} &\multirow{2}{*}{\!\!\cmark\!\!}&\!\!\color{gray}{PixelNeRF}~\cite{yu2021pixelnerf}\!\! &\color{gray}{14.438}&\color{gray}{0.577}&\color{gray}{0.467}&\color{gray}{0.047}&\color{gray}{17.160}&\color{gray}{0.527}&\color{gray}{0.496}&\color{gray}{0.029}\\
          &&\!\!\color{gray}{Du \textit{et al.}}~\cite{du2023learning} &\color{gray}{21.833}&\color{gray}{0.294}&\color{gray}{0.736}&\color{gray}{0.011}&\color{gray}{25.482}&\color{gray}{0.304}&\color{gray}{0.769}&\color{gray}{0.005}\\\cline{2-11}
          &\multirow{3}{*}{\!\!\xmark\!\!}&\!\!DBARF~\cite{chen2023dbarf} &14.789&0.490&0.570&0.033&14.189&0.452&0.537&0.038\\
          &&\!\!FlowCAM~\cite{smith2023flowcam}\!\!&18.242 & 0.597 & 0.455 & 0.023 & 20.116 & 0.477 & 0.585 & 0.016\\
       \cline{3-11}
          &&\!\!Ours &\textbf{19.536}&\textbf{0.398}&\textbf{0.638}&\textbf{0.016}&\textbf{22.440}&\textbf{0.323}&\textbf{0.649}&\textbf{0.010}\\\midrule
        
    \end{tabular}
    }\vspace{-6pt}
    \caption{\textbf{Novel view rendering performance on RealEstate-10K and ACID. } Gray text indicates methods not directly comparable for their use of ground-truth pose at evaluation.} 
    \label{tab:1}\vspace{-10pt}
\end{table}

To quantitatively compute the overlapping regions, 
we employ a pre-trained state-of-the-art dense image matching method~\cite{truong2023pdc} to find the overlapping ratio $o_{12}$ within the image, defined as the ratio of pixels in $I_1$ whose correspondence with pixels in $I_2$ is found with high confidence. We define the overlap between two images to be the intersection over union of two images as $overlap = \frac{1}{o_{12}^{-1} + o_{21}^{-1} - 1}$, and consider images with overlap greater than 0.75 as \textit{Large}, less than 0.5 as \textit{Small}, and the in-between as \textit{Medium}. Finally, the evaluation metrics are computed using synthesized novel view $\hat{I}_t$ and the estimated relative pose between $I_1$ and $I_2$, $\hat{P}_{1\leftarrow 2}$ and those of the ground-truths.    
\subsection{Experimental Results}
\paragraph{Relative Pose Estimation.}
We report quantitative results in Tab.~\ref{tab:2} and the visualization of epipolar lines from the estimated camera poses are shown in Fig.~\ref{fig:epipolar}.  From the results, we observe that our framework significantly outperforms the existing pose-free NeRFs, where they fail to estimate reliable camera pose which can lead to poor view-synthesis quality. Moreover, we observe that  compared to pose estimation methods~\cite{zhang2022relpose,rockwell20228}, our framework achieve  significantly better accuracy, demonstrating the effectiveness of the captured synergy between pose estimation, rendering and image matching. However, it is also notable that PDC-Net+~\cite{truong2023pdc} achieves better performance. This is because PDC-Net was learned with GT correspondences that are obtained using depth information, which indicates further improvements in all our three tasks can be promoted if depth information is incorporated in our framework.\vspace{-12pt}

\begin{table*}[]
    \centering
    \scalebox{0.53}{
    \begin{tabular}{clcccccccccccccccccccc}
    \hlinewd{0.8pt}
        
 &\multirow{4}{*}{\textbf{Components}}  &\multicolumn{5}{c}{\textbf{Avg}}  &\multicolumn{5}{c}{\textbf{Large}}   &\multicolumn{5}{c}{\textbf{Medium}}  &\multicolumn{5}{c}{\textbf{Small}}  
     \\\cmidrule(rl){3-7} \cmidrule(rl){8-12} \cmidrule(rl){13-17}\cmidrule(rl){18-22}
        &&\multirow{2}{*}{PSNR} &\multirow{2}{*}{SSIM} &\multirow{2}{*}{LPIPS} &Mean &Mean. &\multirow{2}{*}{PSNR}&\multirow{2}{*}{SSIM}  &\multirow{2}{*}{LPIPS} &Mean &Mean. &\multirow{2}{*}{PSNR} &\multirow{2}{*}{SSIM} &\multirow{2}{*}{LPIPS} &Mean &Mean&\multirow{2}{*}{PSNR}&\multirow{2}{*}{SSIM}  &\multirow{2}{*}{LPIPS} &Mean &Mean\\
        &&&&&Rot.(\degree)&Trans.(\degree) &&&&Rot.(\degree) &Trans.(\degree) &&&&Rot.(\degree) &Trans.(\degree)&&&&Rot.(\degree) &Trans.(\degree)\\
        \shline
        \textbf{(I)} &Baseline&14.646&0.553&0.513&29.123&52.237 &16.430&0.626&0.406&27.622&71.047&14.246&0.532&0.524&28.636&51.675&13.624&0.522&0.578&30.275&40.164\\
        \textbf{(II)} &+ pose loss  &16.03&0.547&0.485&8.755&62.246&18.872&0.638&0.367&4.303&74.553&16.311&0.548&0.476&8.193&62.510&14.087&0.488&0.565&11.855&54.127\\
        \textbf{(III)} &+ flow head (SSIM loss)&17.393&0.578&0.440&6.737&43.104&17.964&0.588&0.419&4.257&34.584&20.083&0.658&0.329&2.578&54.666&15.439&0.522&0.520&10.325&38.554\\
        \textbf{(IV)} &+ cycle loss&17.899&0.593&0.432&6.675&43.132&20.555&0.662&0.321&2.624&50.541&18.882&0.590&0.412&4.137&34.882&15.821&0.549&0.512&10.210&38.107\\
        \textbf{(V)} &+ aggregation module&18.629&0.611&0.406&5.008&24.769&21.402&0.689&0.294&1.977&31.295&19.219&0.621&0.383&2.951&19.509&16.614&0.556&0.487&7.706&22.363\\ 
        \textbf{(VI)} &+ matching distribution &\textbf{19.536}&\textbf{0.638}&\textbf{0.398}&\textbf{3.610}&\textbf{12.766}&\textbf{22.542}&\textbf{0.724}&\textbf{0.250}&\textbf{1.530}&\textbf{10.187}&\textbf{19.965}&\textbf{0.645}&\textbf{0.343}&\textbf{2.183}&\textbf{11.860}&\textbf{17.153}&\textbf{0.577}&\textbf{0.459}&\textbf{3.610}&\textbf{12.766}\\
       

       \hlinewd{0.8pt}
    \end{tabular}}\vspace{-7pt}
    \caption{\textbf{Component ablations on RealEstate10K}.}\vspace{-3pt}
    \label{tab:component}
\end{table*}

\begin{table*}[t]
\subfloat[\textbf{Fixed Pose and Generalized NeRF}\label{tab:fixed}]{%
\tablestyle{1.0pt}{0.9}
\begin{tabular}{@{}lx{23}x{23}x{23}x{23}x{23}x{23}@{}}
&PSNR  & SSIM &LPIPS & R (\degree) &t (\degree) &t (m) \\
\shline
PDC-Net+. +~\cite{du2023learning}  &18.140  & 0.606&0.366 &\textbf{2.091}&\textbf{8.817}&0.696 \\
Rockwell \textit{et al.}~\cite{rockwell20228} +~\cite{du2023learning}  &14.892  &0.562&0.500  &12.588&90.189&0.524 \\
\textbf{Ours}&\textbf{19.526}&\textbf{0.641}&\textbf{0.312}&2.739&11.362&\textbf{0.290}
\\
\end{tabular}
}\hfill%
\subfloat[\textbf{Pose Supervision }\label{tab:supervision}]{%
\tablestyle{1.0pt}{0.9}
\begin{tabular}{@{}lx{23}x{23}x{23}x{23}x{23}@{}}
&PSNR  & SSIM &LPIPS & R (\degree)&t (\degree)\\
\shline
DBARF~\cite{chen2023dbarf} + pose loss ($\mathcal{L}_\text{pose}$)&{12.998}&{0.468}&{0.566}&{11.82}&{80.66}\\
FlowCAM~\cite{smith2023flowcam} + pose loss ($\mathcal{L}_\text{pose}$)&{18.646}&{0.589}&0.433&{7.505}&{44.347}\\
\textbf{Ours} &\textbf{19.536}&\textbf{0.638}&\textbf{0.398}&\textbf{3.610}&\textbf{12.766}
\end{tabular}
}\hfill

\subfloat[\textbf{Training Strategy}\label{tab:training}]{%
\tablestyle{1.0pt}{0.9}
\begin{tabular}{@{}lx{23}x{23}x{23}x{23}x{23}@{}}
&PSNR  & SSIM &LPIPS & R (\degree)&t (\degree)\\
\shline
Baseline + w/o Teacher Forcing &13.856 &0.502&0.577&31.112&104.490\\
Baseline&14.646&0.553&0.513&29.123&52.237\\
Ours + w/o Teacher Forcing &18.785&0.635&0.415&5.254&40.571\\
\textbf{Ours}&\textbf{19.536}&\textbf{0.638}&\textbf{0.398}&\textbf{3.610}&\textbf{12.766}\\

\end{tabular}
}\hfill
\subfloat[\textbf{The learned representation}\label{tab:representation}]{%
\tablestyle{1.0pt}{0.9}
\begin{tabular}{@{}lx{23}x{23}x{23}x{23}x{23}@{}}
&PSNR  & SSIM &LPIPS & R (\degree)&t (\degree)\\
\shline
 Du \textit{et al.}~\cite{du2023learning} + Noisy Pose ($\sigma$ = 0.025) &18.850&0.618&0.363&2.292&6.171\\
         Du \textit{et al.}~\cite{du2023learning} + GT Pose&21.833&0.736&\textbf{0.294}&-&-\\
         Ours + Noisy Pose ($\sigma$ = 0.025)  & 19.500 &0.633&0.353&2.292&6.171\\
         Ours + GT Pose& \textbf{22.781} &\textbf{0.758}&{0.314}&-&- \\
\end{tabular}
}%
\vspace{-2mm}

\caption{\textbf{More Ablations and insights. See text for details.}}
\label{tab:abl}
\vspace{-3mm}
\end{table*}

\paragraph{Novel View Synthesis.}
Tab.~\ref{tab:1} shows quantitative comparisons, whereas Fig.~\ref{fig:realestate} show qualitative comparisons. From the results, compared to previous pose-free approaches~\cite{smith2023flowcam,chen2023dbarf}, our approach outperforms them all. Note that we also include results from generalized NeRFs~\cite{yu2021pixelnerf,du2023learning} to highlight the complexity and challenge of this task. It’s important to note that while our method may not surpass the state-of-the-art~\cite{du2023learning}, the proximity of our results to it underlines the potential and effectiveness of our approach in  the absence of  camera pose information.

\subsection{Ablation Study and Analysis}
\paragraph{Component ablation.}
In Tab.~\ref{tab:component}, we validate the effectiveness of each component within our framework. The baseline in the first row represents a variant equipped with only the feature backbone, renderer, pose head and image reconstruction loss. We then progressively add each component. From the results, we observe clear improvements on performance for every component, demonstrating that they all contribute to the final performance. A particularly illustrative comparison are (I) vs (II) and (II) vs (III), where simply adding each loss leads to apparent improvements, indicating that the process of finding correspondences, learning 3D geometry through rendering and estimating camera all contribute largely to the performance. However, it is also notable that PDC-Net+. +~\cite{du2023learning} reports lower rotation and translation angular errors. This can be attributed to the use of classical solvers~\cite{fischler1981random, nister2004efficient} that are known to output more precise transformations given a sufficient numbe of correspondences~\cite{Rockwell2024}.\vspace{-12pt}

\paragraph{Will our method be more effective than the combination of readily available models from separate tasks?}
Unless our framework achieves more competitive rendering quality, the practicality of our method will be rather limited. In this analysis, we compare our framework with the variants that adopt two separate methods for camera pose estimation and rendering. The results are reported in Table~\ref{tab:fixed}. Specifically, for the first row, we combine pretrained generalized NeRF~\cite{du2023learning} with an off-the-shelf matching network for pose estimation. The second row shows the outcomes obtained with~\cite{rockwell20228}.  Note that RelPose~\cite{zhang2022relpose} does not predict translations, and we thus leverage~\cite{rockwell20228}.   Summarizing the results, we observe that  our approach outperforms the other variants by large margin, highlighting the importance of capturing the underlying synergy between the tasks and the practicality of the proposed approach.\vspace{-12pt}

\paragraph{Direct pose supervision. }
To assess the impact of direct pose supervision on the rendering and pose estimation performance of existing methods, we explore the potential enhancement of~\cite{smith2023flowcam,chen2023dbarf}
through the integration of direct supervision. For this experiment, we modify them by incorporating the same loss signals as our approach, specifically the geodesic rotation loss and L2 distance loss for translation. The results are reported in Table~\ref{tab:supervision}. 

Although we have found that the use of direct pose supervision aiming to harness benefits from synergistic relationships among different tasks is crucial, when applied to existing frameworks, we have observed only marginal improvements in image quality and a decline in the performance of pose estimation for FlowCAM, while overall decline is obeserved in the performance of DBARF. This outcome is primarily attributed to the architectural design of FlowCAM, where each module operates in a relatively isolated manner without a focus on seamless integration. Conversely, the performance reduction observed in DBARF is multifaceted, with specific causative factors being challenging to pinpoint.  These findings are expounded in the \emph{supp. material} for further discussion. 
In contrast, our proposed framework demonstrates an inherent advantage in harnessing the benefits of pose supervision without requiring further considerations.    \vspace{-12pt}

\paragraph{Ablation study on our training strategy.} As explained in Section~\ref{sec:3.6}, we adopt a special training strategy that conceptually bears similarity to the teacher forcing strategy. In Table~\ref{tab:training}, we validate whether this strategy actually helps. For this experiments, we evaluate two variants that builds upon either \textit{Baseline} or \textit{Ours}: the variant that uses the estimated camera poses at training phase and the other that uses the ground-truth. Comparing the results, we observe clear performance differences between the variants,  demonstrating the effectiveness of the proposed strategy.  \vspace{-12pt}

\paragraph{The learned representation.}
In Table~\ref{tab:representation}, we compare four variants: the first two rows show results from the state-of-the-art generalized NeRF method with ground-truth and noisy poses, respectively, while the next two rows detail results using our framework under the same conditions.  The noisy poses are synthetically perturbed from the ground-truth poses by adding Gaussian noise with $\sigma = 0.025$. From the results, we can observe that that our unified framework's learned representation markedly improves rendering performance, outperforming the current state-of-the-art generalized NeRF method when using ground-truth poses. These results further highlight the importance of jointly learning the three tasks in improving the capabilities of the shared representation.


\section{Conclusion}
In this work, we have presented a novel unified framework that  integrates camera pose estimation, NeRF rendering and correspondence estimation. This approach effectively overcomes the limitations of existing approaches, particularly in scenarios with limited data and complex geometries. Our experimental results, encompassing both indoor and outdoor scenes with only a pair of wide-baseline images, demonstrate the framework's robustness and adaptability in achieving high-quality novel view synthesis and precise camera pose estimation. Extensive ablation studies further validates our choices and highlight the potential of our method to set a new standard in  this task.
\vspace{-10pt}
\section*{Acknowledgement}

This research was supported by the MSIT, Korea (IITP-2023-2020-0-01819, RS-2023-00222280). 
\clearpage

This document includes the following contents: 1) more architectural details of our method, 2) more training and evaluation details of our method and others, 3) distribution of our overlap-based data splitting,  4) more discussions about the experimental results, and  5) additional quantitative and qualitative results for the comparison with other methods and our ablation study.



\section*{A. Architectural Details}
\subsection*{A.1. Feature Aggregation and Cost Filtering}
\begin{figure}
    \centering
    \includegraphics[width=1\linewidth]{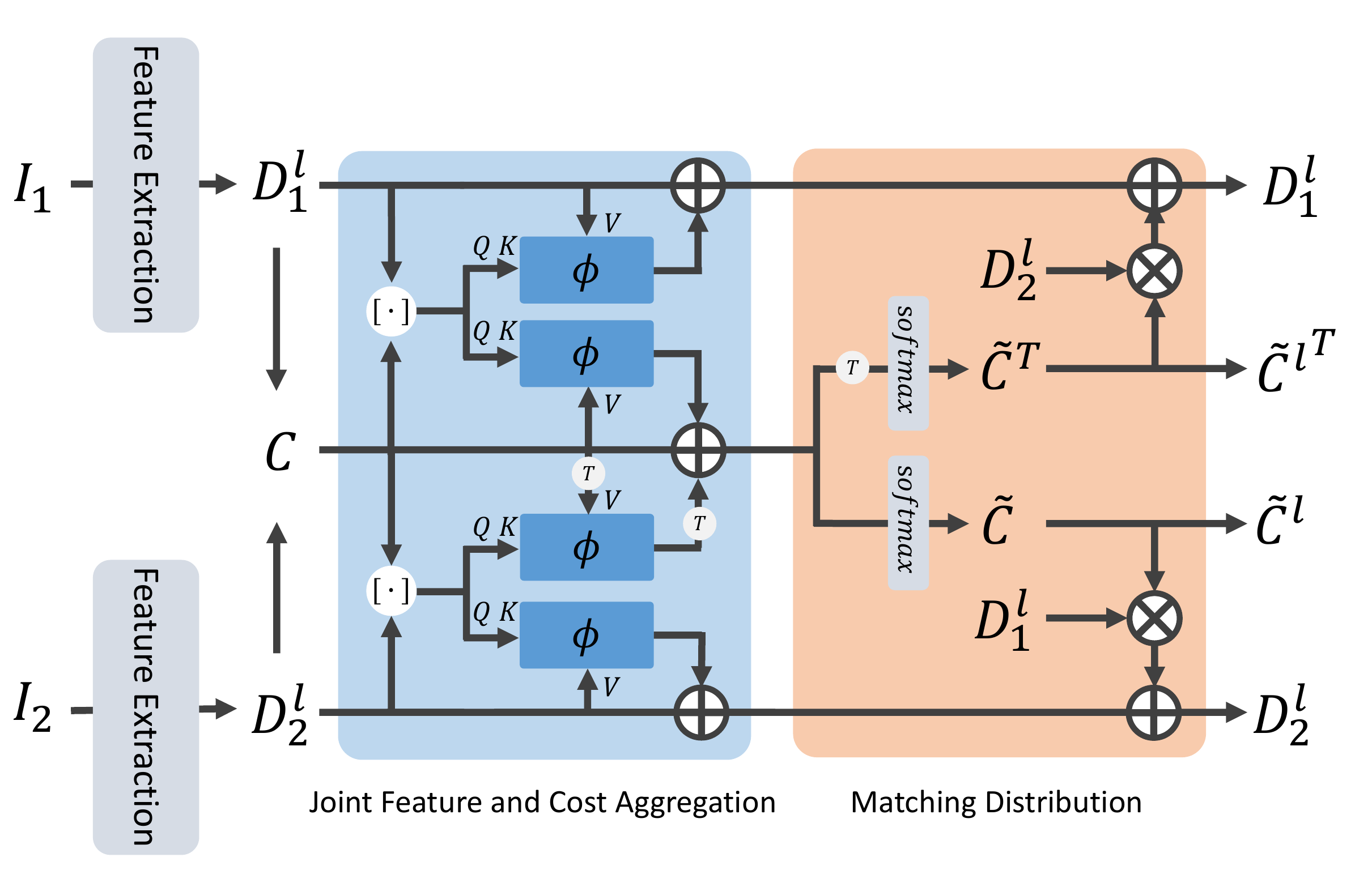}\hfill\\
    \vspace{-5pt}
    \caption{\textbf{Overview of aggregation module.}}
    \label{fig:aggregation}\vspace{-10pt}
\end{figure}

Analogous to existing methods, we  use  attention-based operations for refining both feature and cost volume. We present an overview of the adopted aggregation module in Fig.~\ref{fig:aggregation}. 

\subsection*{A.2. Loss Signals}
\begin{figure}
    \centering
    \includegraphics[width=0.8\linewidth]{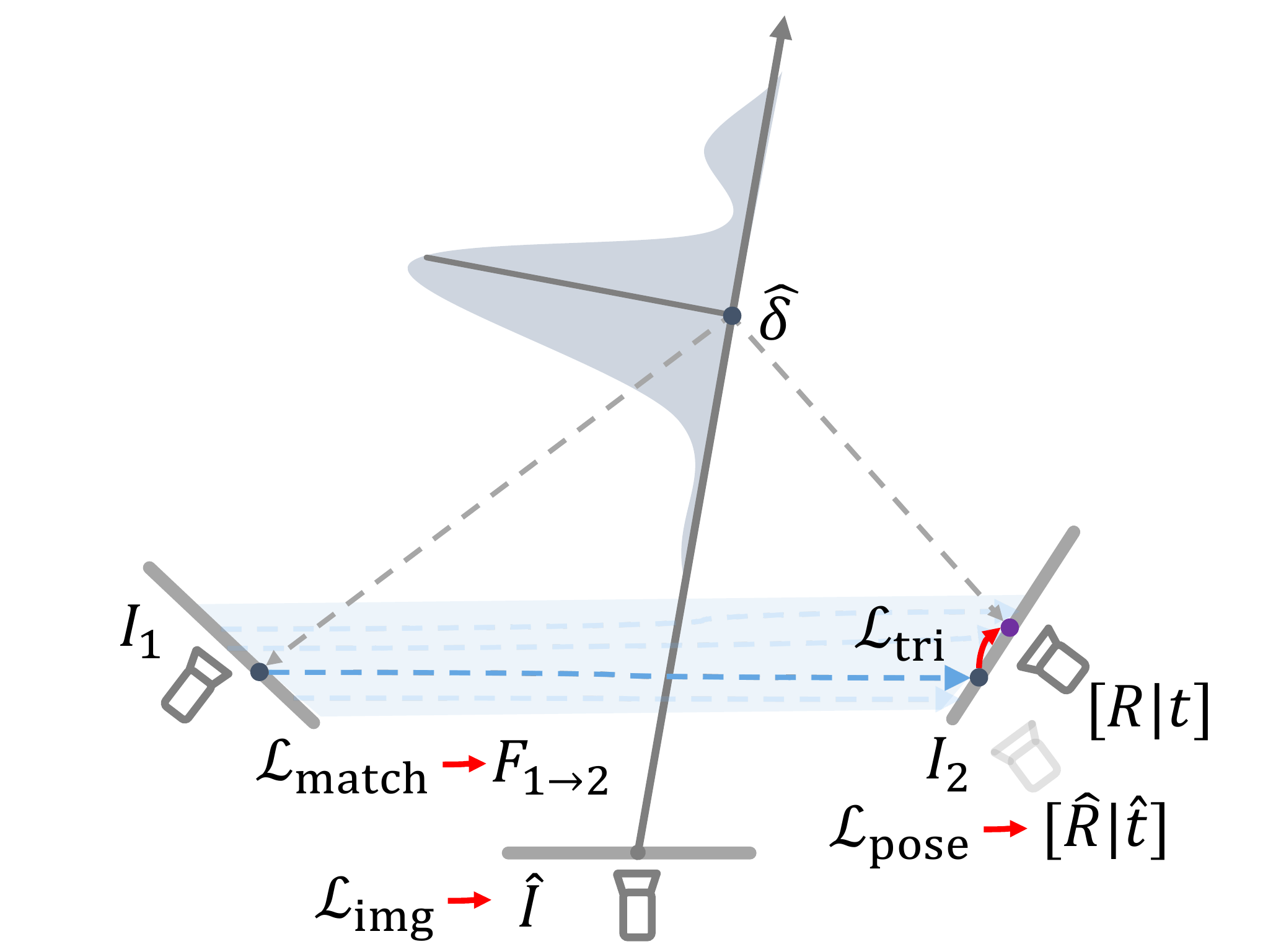}\hfill\\
    \vspace{-5pt}
    \caption{\textbf{Illustration of our training losses.}}
    \label{fig:losses}\vspace{-10pt}
\end{figure}

In Fig.~\ref{fig:losses}, we show an illustration of our training losses. As shown in the figure, the rendering loss is computed between $\hat{I}$ and $I$, pose loss is computed using the estimated camera pose $[\hat{R}|\hat{t}]$, flow loss is computed using the estimated flow $F_{1\rightarrow2}$ and the triplet consistency loss is computed using $\hat{\delta}, [\hat{R}|\hat{t}],$ and $ F_{1\rightarrow2}$.

\section*{B. More Training and Evaluation Details}

\subsection*{B.1. Training Details}
\subsection*{B.1.1. Our Method}
\paragraph{Training Strategy.}
Our training procedure closely resembles that of Du et al.~\cite{du2023learning}, with distinctions in data usage and augmentation. Instead of applying random cropping and flipping, as done by Du et al., we used a subset of datasets without data augmentation. We train the model with $4$ A6000 GPUs for $1\!\!\sim\!\!2$ days, iterating for 50K iterations,  with $192$ rays and $64$ sampled points on epipolar lines. With this configuration, our rendering speed is approximately 0.4 FPS.  

\subsection*{B.1.2. FlowCAM~\cite{smith2023flowcam}}
\paragraph{Architecture.} 
The inference process of FlowCAM is divided into three distinct steps. First, each frame within a video sequence undergoes feature extraction, yielding deep features and backward flows. Subsequently, using the single-view pixelNeRF algorithm to a surface point cloud, representing the anticipated 3D termination point for each pixel.
Next, within each frame, confidence weights are computed among the points. This is achieved by utilizing the RAFT~\cite{teed2020raft} predictions. Finally, these computed confidence weights are fed into a sequence of linear layers to derive the final confidence weights required for solving the Weighted Procrustes formulation~\cite{choy2020deep}, and then the desired views are re-rendered.

\paragraph{Training Strategy.}
For training, we take 256$\times$256 input images.  While RealEstate10K was already trained by the authors and the pretrained weights were available, we verify that the training scheme authors provide can reliably transfer to ACID, we attempted reproducing the results on RealEstate first, for which we were able to reproduce the results close to those reported in the paper. Given this success, we followed the same training procedure used and released by the authors for RealEstate to train on ACID.  We train for 50K iterations with a single A6000 GPU, which takes approximately 1.5 days, with leaving other hyperparameters unchanged.

\subsection*{B.1.3. Rockwell \textit{et al.}~\cite{rockwell20228}}
\paragraph{Architecture.} In their architecture, there are three main components: Image encoder, ViT layer and Essential Matrix Module followed by MLPs. Taking 256$\times$256 input image pairs as inputs, the image is resized to 224$\times$224, and then the model first extracts deep features via vanilla resnet-18. Then the feature maps from the coarsest layer are fed to ViT-Tiny for self-attention operations. Subsequently, these feature maps are fed to the Essential Matrix Module, which performs the cross-attention that emulates the 8-point algorithm, and finally, the output is reshaped and fed to the pose regression MLPs.   

\paragraph{Training Strategy.}
For training, we follow the same procedure and adopt the default hyperparameters used in the training scripts,  as provided in the official github repository that the authors provide for both RealEstate10K and ACID. We use the same data sampling strategy as the one we used to train our model. Specifically, for each scene consisting of a video sequence, we use the first and the last frame as the input images, and the ground-truth relative pose for supervision is computed between them. We trained the network for a total of 120K iterations with batch size set to 32 using a single A6000 GPU.

\subsection*{B.1.3. RelPose~\cite{zhang2022relpose}}
\paragraph{Architecture.} Relpose inference is divided into two steps. A pairwise pose prediction step is followed by a joint reasoning step of multiple pairwise estimated relative poses. By taking a set of images as input, they first group all possible pairs of images to estimate all the pairwise relative poses between images. Leveraging an energy-based model, the estimated pairwise relative poses recover a probability distribution over conditional relative rotations where the condition is given as the uniformly sampled relative pose $R \in \mathrm{SO(3)}$. Estimated poses are further refined in the joint reasoning step by inducing a joint likelihood over the camera transformations across multiple images and iteratively improving an initial estimate by maximizing this likelihood.

\paragraph{Training Strategy.} For training, we follow the same training strategy as \cite{du2023learning} and ours, since we aim to compare the performance of relative pose estimation given stereo pairs. However, when using only stereo pairs as input, the joint reasoning step cannot be done as there is only one estimated pose. To make a fair comparison, we increased the number of uniformly sampled relative pose $R \in \mathrm{SO(3)}$ from $N=36864$ to $N=250000$, which is the number of queries used in the second stage of the framework. The training was done for 400K iterations of batch size set to 64, using four A6000 GPUs.

\subsection*{B.1.4. DBARF~\cite{chen2023dbarf}}
\paragraph{Architecture.} The architecture of DBARF consists of three components: an image encoder, a Pose and Depth Estimation Module, and a Renderer for novel view synthesis. By selecting a target image and nearby images from a scene graph, the ResNet-like~\cite{he2016deep} image encoder first extracts a feature map used for the subsequent steps. The feature maps of the nearby images are then warped to the target view using the currently estimated camera poses and depths to construct a local cost map for pose and depth estimation done with training a recurrent GRU. The estimated pose is then used as an input of the Renderer, where they use the IBRNet~\cite{wang2021ibrnet} to render novel views. To enable robust optimization of both the Pose and Depth Estimation Module and the Renderer, they adopt a staged training strategy of dividing the overall training process into three steps: training only the Pose and Depth Estimation Module, training only the Renderer, and jointly training the two components.
\begin{figure*}
    \centering
    \includegraphics[width=1.0\linewidth]{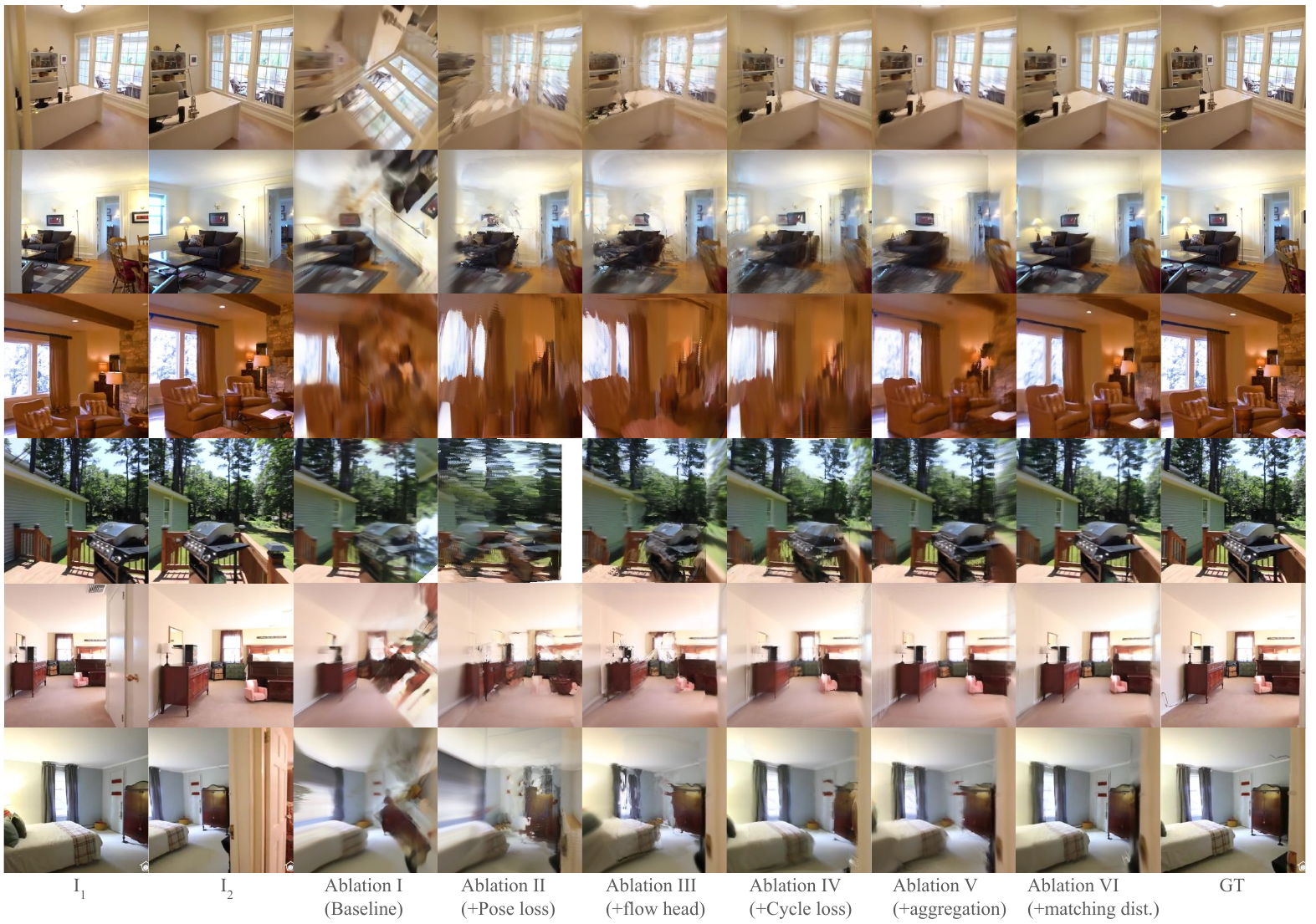}\vspace{-5pt}
    \caption{\textbf{Qualitative results for component ablation study.} Consistent with the quantitative results (Table 3 of the main paper), each variant exhibits apparent differences in qualitative comparisons and shows the efficacy of our designed components.}
    \label{fig:ablation}
\end{figure*}
\paragraph{Training Strategy.} For training, we first take 256$\times$256 input images and then resize them to 224$\times$224. As there are no provided weights for DBARF on RealEstate10K and ACID, we trained the network from scratch following the process provided by the authors. For both datasets, we selected six nearby views of the target view during training by selecting three frames before the target frame with a 10, 20, and 30 frame difference each and three frames after the target frame with a 10, 20, and 30 frame difference. We trained the network for a total of 200K iterations, where the three stages of their proposed staged training were repeated every 10K steps. The training was done with a single A6000 GPU.

\begin{figure}
  \centering
    \subfloat[RealEstate-10K]
    {\includegraphics[width=0.495\linewidth]{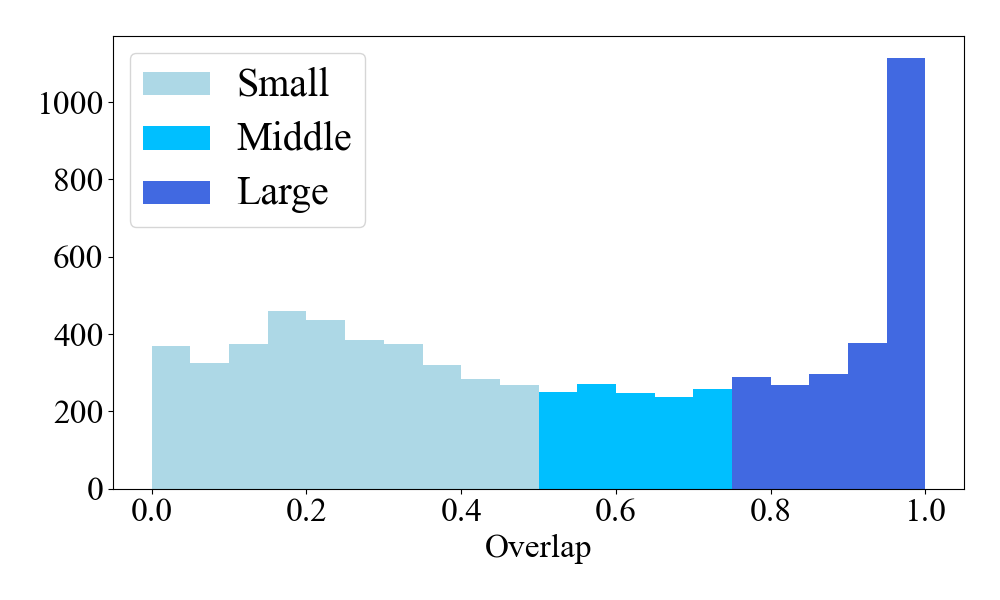}}\hfill
    \subfloat[ACID]
    {\includegraphics[width=0.495\linewidth]{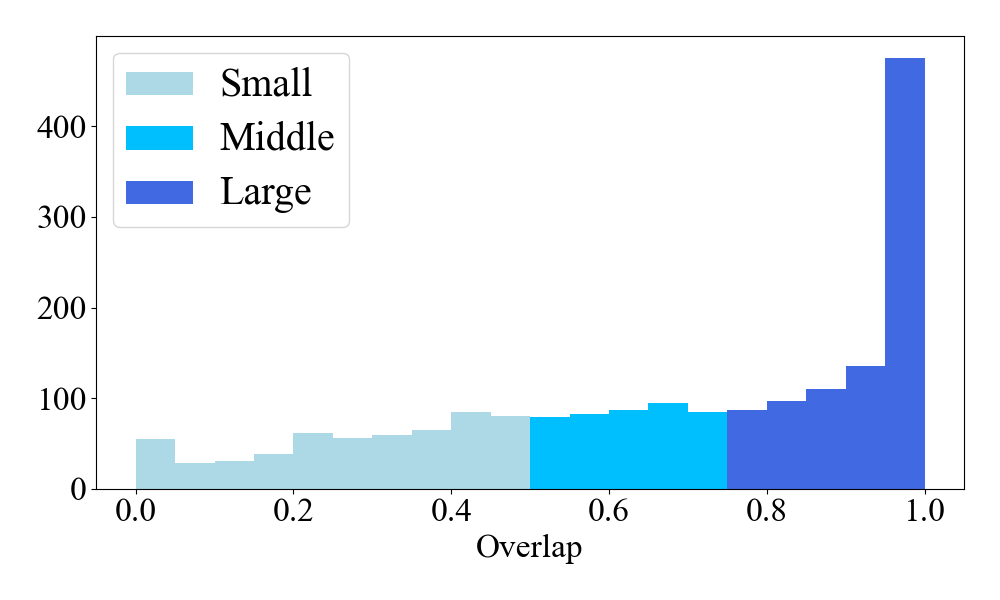}}\hfill\\
\vspace{-5pt}
    \caption{\textbf{Distribution of the data splits.} }
    \vspace{-15pt}
  \label{fig:distribution}
\end{figure}


\subsection*{B.2. Evaluation Details}
\paragraph{Differences of the evaluation strategy in original work of DBARF and FlowCAM and our adopted evaluation strategy.}

In the original evaluation strategy of DBARF, given a sequence of frames, DBARF picks an abitrary view, treating it as a target view, and selects nearby views, considering them as context images. Subsequently, pairwise pose estimation and depth estimation are performed between the target image and each of the context images. The estimated values are then fed to GeNeRF for rendering and evaluation. In our evaluation approach, we assume that only two context images and a relative camera pose to the target view are provided. This means that in contrast to the original evaluation setting, where target view is accessible for depth and pose estimation, our evaluation setting does not employ the target view for the model, but only accessible for metric computation.

In the evaluation setting of FlowCAM, FlowCAM first feeds all the frames to single-view PixelNeRF, and they are used for warping by an off-the-shelf optical flow network to output matching confidence that will be used for Weighted Procrustes formulation. Similar to DBARF, the target views are accessed for pose estimation in this evaluation setting. In our evaluation setting, we substitute the estimated poses with ground-truth poses except for the relative pose between $I_1$ and $I_2$, which is their prediction and remains unchanged.  

\paragraph{Fixed Pose and Generalized NeRF.} In Table 4 of the main paper, we conducted additional ablation study and analysis. While all other experiments follow the same evaluation protocol, some subtle changes are made for (a). Specifically, as PDC-Net encountered some cases where RANSAC failed to converge due to extremely small overlapping regions with barely any correspondence, we exclude such cases during metric calculations. For a fair comparison, for the other variants, \ie, Rockwell \textit{et al.}+\cite{du2023learning} and ours, we also disregard those scenes, which leads to different results compared to our main table.

\section*{C. Data Split Details}
 We provide the statistics of each overlap-based data split of RealEstate10K and ACID in Fig.~\ref{fig:distribution}.

\section*{D. More Discussions}

\paragraph{Pose Supervision.}
In this section, we discuss the results we obtained from the experiment that combines direct pose supervision to FlowCAM and DBARF.
For FlowCAM, we hypothesize that the disjoint design among each modules, \textit{i.e.,} pose and renderer, and the lack of coherence among the modules results in limited overall benefits. Because off-the-shelf flow model is left frozen for inference, it will inevitably risk finding inaccurate poses when there are only minimal overlapping regions between image pairs, which underscores the effectiveness of our proposed approach that emphasizes the meticulous unification of various tasks to effectively harness their synergistic potential. 
For DBARF, contrary to initial expectations, adding direct pose supervision results in a decline in their pose estimation performance despite our meticulous efforts to optimize the hyperparameters and conduct multiple trials. 
We hypothesize that the performance degradation could be attributed to their original training strategy, which only assumes image sequences with small viewpoint changes and could have influenced overall performance largely.
Another potential reason is that their incorporation of pose estimation and the rendering requires non-trivial implementation considerations to obtain a boosted synergy, or the confidence score produced by the off-the-shelf model might've made a disparity between the updated renderer and the optical flow prediction. 
Despite the challenges we faced and the hypothesis we made, we leave this exploration as future work since such investigation is beyond the scope of this paper.

\section*{E. More Results}
\subsection*{E.1. Absolute Translation Error with Scales}
Table~\ref{tab:scale} presents the results of translation estimation evaluated with both absolute error in meters and angular difference in degrees. Note that, as mentioned in the main paper, evaluating absolute error requires assessing the models' ability to gauge scales via, \eg, object recognition, since translation scale is theoretically indeterminable in two-view geometry. This could potentially result in erroneous interpretations regarding the models' proficiency in estimating relative camera pose from two views.

\begin{table*}[]
\centering
\scalebox{0.685}{
    \begin{tabular}{l|c|l|ccc|ccc|ccc|ccc}
    \toprule
         \multirow{3}{*}{Overlap}&\multirow{3}{*}{Task}&\multirow{3}{*}{Method} & \multicolumn{6}{c|}{RealEstate-10K} &\multicolumn{6}{c}{ACID} \\\cline{4-15}
         && &\multicolumn{3}{c}{Translation} &\multicolumn{3}{c|}{Translation} &\multicolumn{3}{c}{Translation} &\multicolumn{3}{c}{Translation}\\
        & & &\!\!Avg(m)$\downarrow$\!\!& \!\!Med(m)$\downarrow$\!\! &\!\!STD(m)$\downarrow$\!\!
         &\!\!Avg(\degree)$\downarrow$\!\!& \!\!Med(\degree)$\downarrow$\!\!&\!\!STD(\degree)$\downarrow$\!\!& \!\!Avg(m)$\downarrow$\!\!& \!\!Med(m)$\downarrow$\!\! &\!\!STD(m)$\downarrow$\!\!
         &\!\!Avg(\degree)$\downarrow$\!\!& \!\!Med(\degree)$\downarrow$\!\! &\!\!STD(\degree)$\downarrow$\!\!\\\midrule 
         \multirow{7}{*}{Small} &\multirow{2}{*}{{Matching}}&\color{gray}{SP+SG}~\cite{detone2018superpoint,sarlin2020superglue,fischler1981random}&\color{gray}{0.973}&\color{gray}{0.759}&\color{gray}{0.840}&\color{gray}{12.549}&\color{gray}{4.638}&\color{gray}{23.048}&\color{gray}{0.979}&\color{gray}{0.661}&\color{gray}{1.094}&\color{gray}{22.214}&\color{gray}{7.526}&\color{gray}{33.719}\\
          &&\color{gray}{PDC-Net+}~\cite{fischler1981random,truong2023pdc} &\color{gray}{0.696}&\color{gray}{0.597}&\color{gray}{0.591}&\color{gray}{6.913}&\color{gray}{2.752}&\color{gray}{15.558}&\color{gray}{0.667}&\color{gray}{0.573}&\color{gray}{0.714}&\color{gray}{15.664}&\color{gray}{4.215}&\color{gray}{29.640}\\\cline{2-15}
         &\multirow{2}{*}{Pose Estimation}&Rockwell \textit{et al.}~\cite{rockwell20228} &1.692&1.459&1.119&91.455&91.499&56.872&1.576&1.057&3.557&{88.421}&{88.958}&{36.212}\\
         &&RelPose~\cite{zhang2022relpose}&-&-&-&-&-&-&-&-&-&-&-&-\\\cline{2-15}
         &\multirow{3}{*}{Pose-Free NeRF}&DBARF~\cite{chen2023dbarf}&2.782&2.549&1.803&126.282&140.358&43.691&2.134&1.187&6.959&95.149&99.490&47.576\\
         &&FlowCAM*~\cite{smith2023flowcam}&{1.543}&{1.400}&{0.901}&{87.119}&{58.245}&{26.895}&{0.732}&{0.487}&{0.810}&{92.130}&{85.846}&{40.821}\\\cline{3-15}
         &&Ours&\textbf{0.532}&\textbf{0.353}&\textbf{0.642}&\textbf{11.862}&\textbf{5.344}&\textbf{21.080}&\textbf{0.378}&\textbf{0.171}&\textbf{0.533}&\textbf{23.689}&\textbf{11.289}&\textbf{30.391}
         \\\midrule
         \multirow{7}{*}{Medium} &\multirow{2}{*}{{Matching}}&\color{gray}{SP+SG}~\cite{detone2018superpoint,sarlin2020superglue,fischler1981random}&\color{gray}{0.390}&\color{gray}{0.344}&\color{gray}{0.261}&\color{gray}{9.295}&\color{gray}{3.279}&\color{gray}{20.456}&\color{gray}{0.528}&\color{gray}{0.466}&\color{gray}{0.431}&\color{gray}{16.455}&\color{gray}{5.426}&\color{gray}{29.035}\\
         &  &\color{gray}{PDC-Net+}~\cite{fischler1981random,truong2023pdc} &\color{gray}{0.360}&\color{gray}{0.322}&\color{gray}{0.253}&\color{gray}{6.667}&\color{gray}{2.262}&\color{gray}{18.247}&\color{gray}{0.612}&\color{gray}{0.563}&\color{gray}{0.482}&\color{gray}{14.940}&\color{gray}{4.301}&\color{gray}{27.379}\\\cline{2-15}
         &\multirow{2}{*}{Pose Estimation}&Rockwell \textit{et al.}~\cite{rockwell20228} &0.842&0.705&{0.581}&82.478&82.920&55.094&0.713&0.554&{0.649}&90.555&90.799&51.469\\
         &&RelPose~\cite{zhang2022relpose}&-&-&-&-&-&-&-&-&-&-&-&-\\\cline{2-15}
         &\multirow{3}{*}{Pose-Free NeRF}&DBARF~\cite{chen2023dbarf}&0.816&0.574&0.782&79.402&75.408&54.485&0.772&0.473&0.931&{77.324}&{77.291}&49.735\\
         &&FlowCAM*~\cite{smith2023flowcam}&{0.563}&{0.510}&0.384&{42.287}&{41.594}&{24.862}&{0.654}&{0.447}&0.768&95.444&87.308&{43.198}\\\cline{3-15}
         &&Ours&\textbf{0.203}&\textbf{0.150}&\textbf{0.178}&\textbf{10.187}&\textbf{5.749}&\textbf{15.801}&\textbf{0.324}&\textbf{0.133}&\textbf{0.615}&\textbf{21.401}&\textbf{10.656}&\textbf{28.243}\\\bottomrule
          \multirow{7}{*}{Large} &\multirow{2}{*}{{Matching}}&\color{gray}{SP+SG}~\cite{detone2018superpoint,sarlin2020superglue,fischler1981random}&\color{gray}{0.612}&\color{gray}{0.665}&\color{gray}{0.202}&\color{gray}{21.415}&\color{gray}{7.190}&\color{gray}{34.044}&\color{gray}{0.619}&\color{gray}{0.641}&\color{gray}{0.260}&\color{gray}{22.018}&\color{gray}{7.309}&\color{gray}{33.775}\\
           &&\color{gray}{PDC-Net+}~\cite{fischler1981random,truong2023pdc} &\color{gray}{0.601}&\color{gray}{0.659}&\color{gray}{0.200}&\color{gray}{16.567}&\color{gray}{5.447}&\color{gray}{29.883}&\color{gray}{0.707}&\color{gray}{0.606}&\color{gray}{0.882}&\color{gray}{18.447}&\color{gray}{4.357}&\color{gray}{35.564}\\\cline{2-15}
         &\multirow{2}{*}{Pose Estimation} &Rockwell \textit{et al.}~\cite{rockwell20228} &0.468&0.363&0.377&91.851&88.923&57.444&0.431&0.304&0.457&86.580&87.559&50.369\\
         &&RelPose~\cite{zhang2022relpose}&-&-&-&-&-&-&-&-&-&-&-&-\\\cline{2-15}
         &\multirow{3}{*}{Pose-Free NeRF}&DBARF~\cite{chen2023dbarf}&0.217&0.098&{0.318}&50.094&33.959&43.659& \textbf{0.281}&\textbf{0.111}&{0.488}  & {54.523} & {38.829} & 45.453\\
         &&FlowCAM*~\cite{smith2023flowcam}&{0.179}&{0.107}&0.197&{34.472}&{27.791}&{30.615}&0.778&0.442&2.789&97.392&89.359&{43.777}\\\cline{3-15}
         &&Ours&\textbf{0.095}&\textbf{0.067}&\textbf{0.102}&\textbf{15.544}&\textbf{7.907}&\textbf{24.626}&{0.456}&{0.146}&\textbf{0.276}&\textbf{22.935}&\textbf{10.588}&\textbf{30.974}\\\bottomrule
         \multirow{7}{*}{\textit{Avg}} &\multirow{2}{*}{{Matching}}&\color{gray}{SP+SG}~\cite{detone2018superpoint,sarlin2020superglue,fischler1981random} &\color{gray}{0.749}&\color{gray}{0.629}&\color{gray}{0.654}&\color{gray}{14.887}&\color{gray}{5.058}&\color{gray}{27.238}&\color{gray}{0.703}&\color{gray}{0.610}&\color{gray}{0.676}&\color{gray}{20.802}&\color{gray}{6.878}&\color{gray}{32.834}\\
        & &\color{gray}{PDC-Net+}~\cite{fischler1981random,truong2023pdc}&\color{gray}{0.696}&\color{gray}{0.0.597}&\color{gray}{0.591}&\color{gray}{10.100}&\color{gray}{3.243}&\color{gray}{22.317}&\color{gray}{0.671}&\color{gray}{0.587}&\color{gray}{0.744}&\color{gray}{16.461}&\color{gray}{4.292}&\color{gray}{31.391}\\\cline{2-15}
         &\multirow{2}{*}{Pose Estimation}&Rockwell \textit{et al.}~\cite{rockwell20228} &1.145&0.821&1.022&90.115&88.648&40.948&0.833&0.500&2.041&88.433&88.961&{36.197}\\
         &&RelPose~\cite{zhang2022relpose}&-&-&-&-&-&-&-&-&-&-&-&-\\\cline{2-15}
         &\multirow{3}{*}{Pose-Free NeRF}&DBARF~\cite{chen2023dbarf}&1.603&0.930&1.787&93.300&102.467&57.290& 0.939&0.366&{3.901}  & {71.711} & {68.892} & 50.277\\
         &&FlowCAM*~\cite{smith2023flowcam}&{0.916}&{0.647}&{0.913}&{50.659}&{46.281}&{52.321}&{1.665}&{1.538}&{0.748}&95.405&88.133&42.849\\\cline{3-15}
         &&Ours&\textbf{0.332}&\textbf{0.177}&\textbf{0.506}&\textbf{12.766}&\textbf{7.534}&\textbf{15.510}&\textbf{0.404}&\textbf{0.150}&\textbf{0.197}&\textbf{22.809}&\textbf{14.502}&\textbf{21.572}\\\bottomrule
    \end{tabular}
    }\vspace{-6pt}
    \caption{Translation estimation performance evaluated with both absolute error (in meters) and angular error (in degrees).  Note that since translation scale is theoretically indeterminable in two-view geometry, evaluating absolute error requires assessing the models' ability to gauge scales via, \eg, object recognition. This could potentially result in erroneous interpretations regarding the models' proficiency in estimating relative pose from two views.  *: FlowCAM~\cite{smith2023flowcam} results have been updated to rectify an error in the numerical values 
originally presented.
    }\vspace{-9pt}
    \label{tab:scale}
\end{table*}

\subsection*{E.2. More Qualitative Results}

\begin{figure*}
    \centering
    \includegraphics[width=0.6\linewidth]{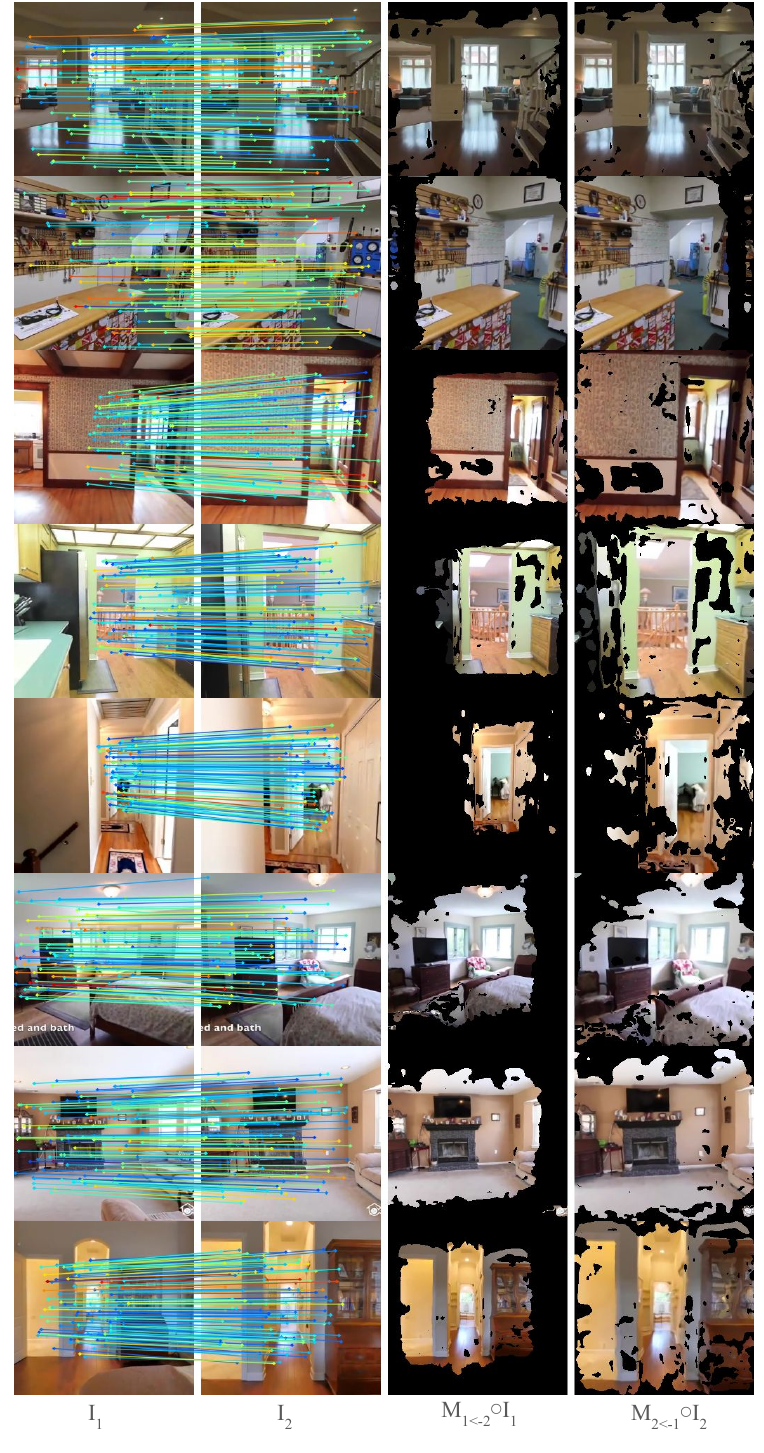}
    \vspace{-10pt}
    \caption{\textbf{Randomly selected correspondences and confident regions. } For each pair of images, we visualize a set of randomly selected correspondences (left), and from the complete set of correspondences, and those with confidence score of higher than a threshold $\tau$ are shown as visible (right).   }
    \label{fig:match_supple}
\end{figure*}

Fig.~\ref{fig:match_supple} shows the correspondences built by our method and the overlapping regions characterized by high confidence scores. As we can see, our method can detect matching points robustly across different scenarios.

\begin{figure*}
    \centering
    \includegraphics[width=0.6\linewidth]{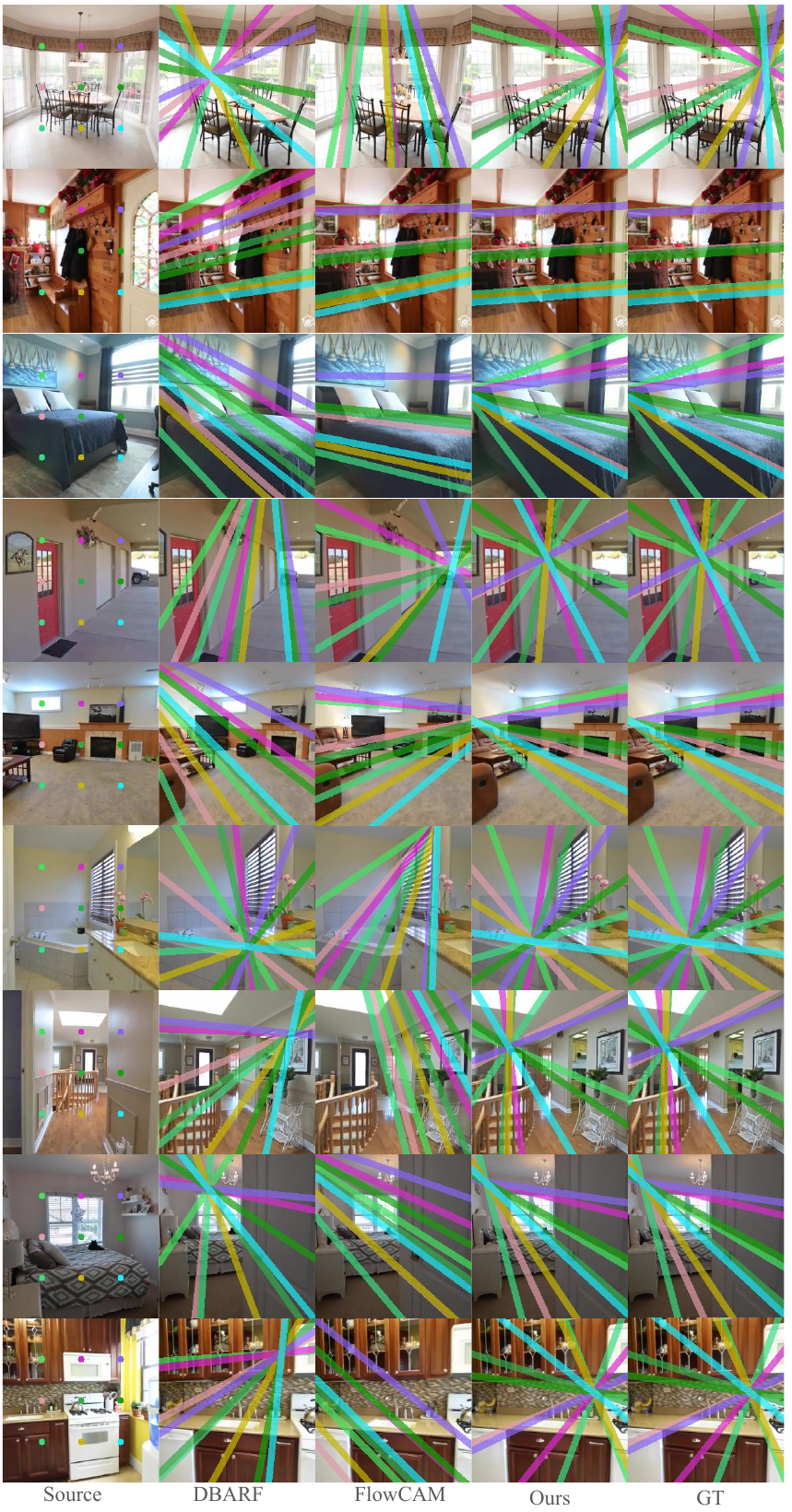}\vspace{-5pt}
    \caption{\textbf{Comparisons of visualized epipolar lines.} }
    \label{fig:epipolar_supple}
\end{figure*}

Fig.~\ref{fig:epipolar_supple} visualizes the epipolar lines with the relative poses estimated by different methods.  Visually inspected, our method yields more accurate results especially on challenging cases with small overlap.

\begin{figure*}
    \centering
    \includegraphics[width=1\linewidth]{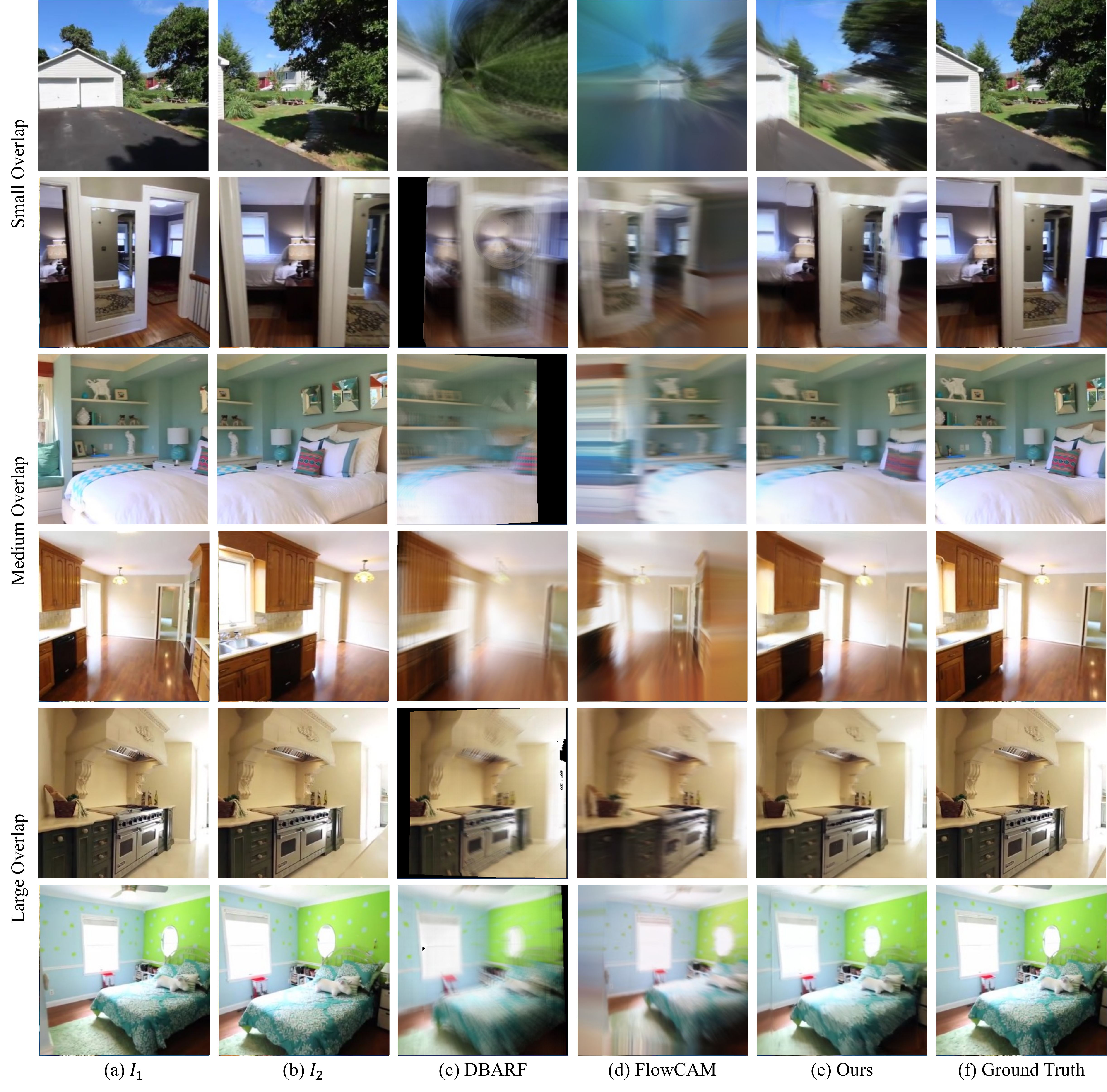}\vspace{-5pt}
    \caption{\textbf{Qualitative comparison on RealEstate10K.} }
    \label{fig:qual_realestate_supple}
\end{figure*}

\begin{figure*}
    \centering
    \includegraphics[width=1\linewidth]{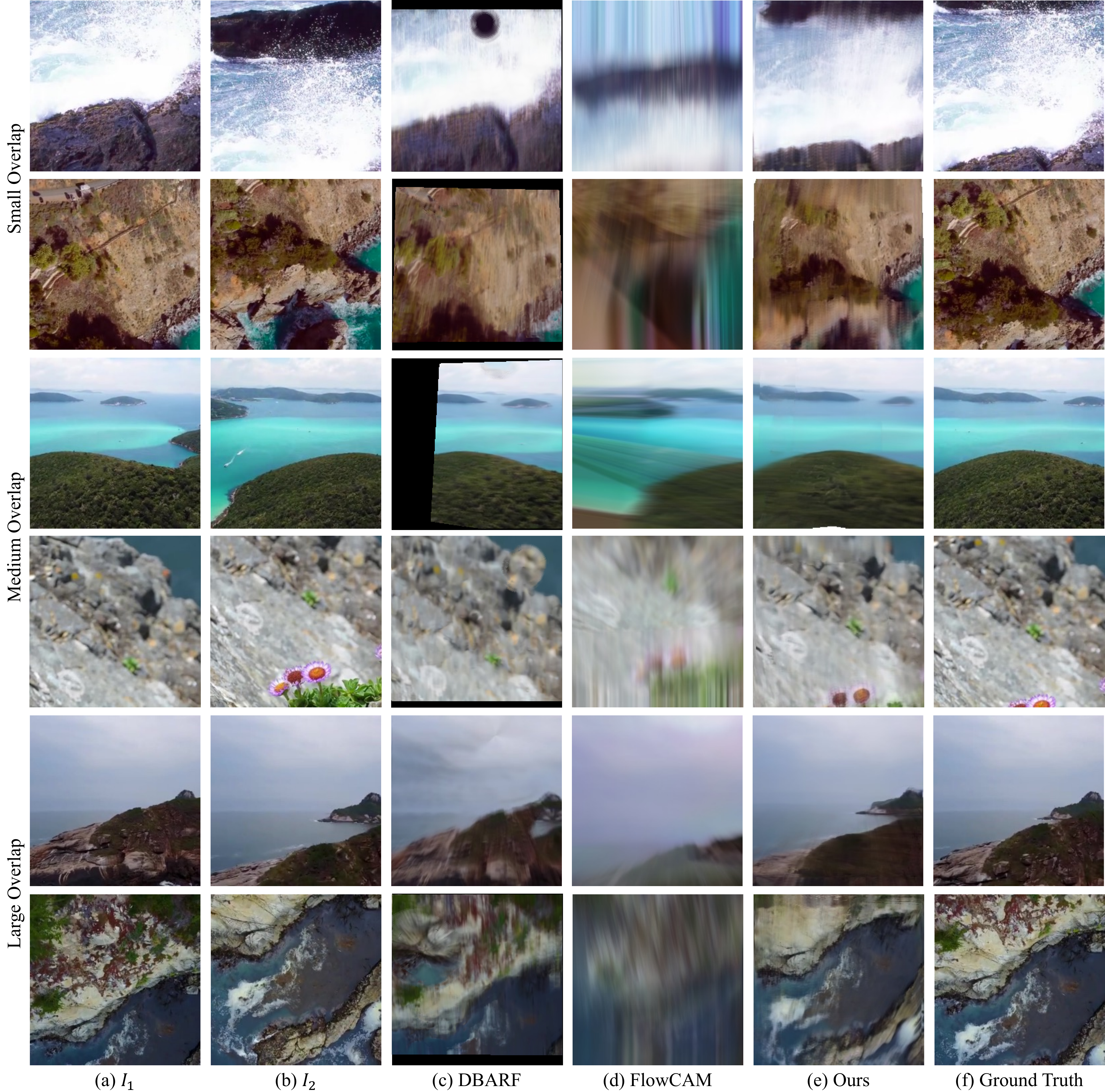}\vspace{-5pt}
    \caption{\textbf{Qualitative comparison on ACID.} }
    \label{fig:qual_acid_supple}
\end{figure*}

Fig.~\ref{fig:qual_realestate_supple} and Fig.~\ref{fig:qual_acid_supple} present more novel view rendering results of different methods. On both datasets, our method yields outcomes that are sharper and more geometrically accurate.


\subsection*{E.3. Qualitative Results for Ablation Study}
In Fig.~\ref{fig:ablation}, we provide qualitative comparisons for each variant introduced for the component ablation study. Consistent with the quantitative results, each variant exhibits apparent differences in qualitative comparisons as well.

\clearpage
\clearpage
\clearpage
\clearpage
\clearpage



\clearpage


\clearpage
{
    \small
    \bibliographystyle{ieeenat_fullname}
    \bibliography{main}
}


\end{document}